\documentclass[10pt,twocolumn,letterpaper]{article}

\usepackage{iccv}
\usepackage{times}
\usepackage{epsfig}
\usepackage{graphicx}
\usepackage{float}
\usepackage{subfig}
\usepackage{amsmath}
\usepackage{amssymb}
\usepackage{booktabs}
\usepackage[accsupp]{axessibility}  
\usepackage[ruled,vlined]{algorithm2e}
\usepackage{multirow}

\usepackage[breaklinks=true,colorlinks,bookmarks=false]{hyperref}

\iccvfinalcopy 

\def\iccvPaperID{1351} 

\ificcvfinal\fi

\begin{document}

\title{Class Prior-Free Positive-Unlabeled Learning with Taylor Variational Loss for\\ Hyperspectral Remote Sensing Imagery}

\author{Hengwei Zhao\qquad Xinyu Wang\thanks{Corresponding author.}\qquad Jingtao Li\qquad Yanfei Zhong\\
Wuhan University, Wuhan, China\\
{\tt\small \{whu\_zhaohw, wangxinyu, JingtaoLi, zhongyanfei\}@whu.edu.cn}
}

\maketitle
\ificcvfinal\thispagestyle{empty}\fi

\begin{abstract}
Positive-unlabeled learning (PU learning) in hyperspectral remote sensing imagery (HSI) is aimed at learning a binary classifier from positive and unlabeled data, which has broad prospects in various earth vision applications.
However, when PU learning meets limited labeled HSI, the unlabeled data may dominate the optimization process, which makes the neural networks overfit the unlabeled data.
In this paper, a Taylor variational loss is proposed for HSI PU learning, which reduces the weight of the gradient of the unlabeled data by Taylor series expansion to enable the network to find a balance between overfitting and underfitting.
In addition, the self-calibrated optimization strategy is designed to stabilize the training process.
Experiments on 7 benchmark datasets (21 tasks in total) validate the effectiveness of the proposed method.
Code is at: \url{https://github.com/Hengwei-Zhao96/T-HOneCls}.
\end{abstract}

\section{Introduction}\label{sec:introduction}
Positive-unlabeled learning is aimed at learning a binary classifier from positive and unlabeled data~\cite{nnpu,PAN,vPU}.
Due to the lack of negative samples, PU learning is a challenging task, but play an important role in machine learning applications, including product recommendation~\cite{10.5555/3045118.3045378}, deceptive reviews detection~\cite{ren-etal-2014-positive}, and medical diagnosis~\cite{2012Positive}.

PU learning in HSI is a powerful tool for environmental monitoring~\cite{ZHAO2022328,LI2022102947}.
For example, when mapping the invasive species in complex forestry, PU learning only needs positive labels
of invasive species; however, traditional hyperspectral classification~\cite{7967742,8737729,FPGA} requires the various negative classes to be labeled to obtain a discriminate boundary, which is labor-intensive, even impossible, to investigate the negative objects and annotate them in high species richness areas~\cite{ZHAO2022328}.

Few releated works have focused on PU learning in HSI.
Compared to other tasks, the training data size in HSI is much smaller~\cite{7882742}, and the deep models are more likely to be over-fitting and susceptible to unalabeled data.
These characteristics make hyperspectral PU learning a more challenging task.

PU learning methods can be divided into two categories, according to whether the class prior ($\pi_p$, i.e., the proportion of positive data) is assumed to be known.
(1) Due to the limited supervision information from PU data, most studies assume that the class prior is available~\cite{ZHAO2022328,LI2022102947}, but in reality, the class prior is hard to be estimated accurately, especially for HSIs, due to the severe inter-class similarity and intra-class variation.
(2) Class prior-free PU learning is a recent research focus of the machine learning community~\cite{vPU,PAN}, where variational principle-based PU learning~\cite{vPU} is one of the state-of-the-art in theory.
It approximates the positive distribution by optimizing the posterior probability, i.e., the classifier, and does not require knowing the class prior.
However, the unlabeled data may dominate the optimization process, which makes it difficult for neural networks to find a balance between the underfitting and overfitting of positive data, especially when the variational principle meets limited labeled HSI data (discussed later in Section~\ref{sec:algorithm} in detail).

In this paper, a Taylor series expansion-based variational framework---\emph{T-HOneCls}---is proposed to solve the limited labeled hyperspectral PU learning problem without class prior.
The contributions of this paper are summarized as follows:
\begin{itemize}
  \item A novel insight is proposed in terms of the dynamic change of the loss, which demonstrates that the unlabeled data dominating the training process is the bottleneck of the variational principle-based classifier.
  \item \emph{Taylor variational loss} is proposed to tackle the problem of PU learning without a class prior, which reduces the weight of the gradient of the unlabeled data and simultaneously satisfy the variational principle by Taylor series expansion, to alleviate the problem of unlabeled data dominating the training process.
  \item \emph{Self-calibrated optimization} is proposed to take advantage of the supervisory signals from the network 
itself to stabilize the training process and alleviate the potential over-fitting problem  caused by limited labeled data with a large pool of unlabeled data.
  \item Extensive experiments are conducted on 7 benchmark datasets, including 5 hyperspectral datasets (19 tasks in total), CIFAR-10 and STL-10, where the proposed method outperforms other state-of-the-art methods in most cases.
\end{itemize}
\section{Related Works}\label{sec:releated_works}
\paragraph{Deep Learning Based Classification for HSI}
The methods of HSI classification can be divided into patch-based framework and patch-free framework~\cite{FPGA}.
The patch-based methods aim to model a mapping function $f_{pb}: R^{S \times S} \rightarrow R$, and first extract the pixels to be classified and their surrounding pixels to build patches with the size $S \times S$, and then use these patches and labels to train a neural network.
Different neural networks can be used to model $f_{pb}$~\cite{6844831,8356713,8662780,8661744}.
The patch-free frameworks aim to model a mapping function $f_{pf}: R^{H \times W} \rightarrow R^{H \times W}$ by a fully convolutional neural network~\cite{7967742,8737729,FPGA}, and due to the avoidance of redundant computation in patches, the inference time of the patch-free frameworks is improved by hundreds of times~\cite{FPGA}.

Differing from the above supervised classification methods, which both need positive and negative data, the method proposed in this paper focuses on weakly supervised PU learning and only requires positive data to be labeled.

\paragraph{Positive and Unlabeled Learning}
Early studies focused on the two-step heuristic approach~\cite{FOODY20061,10.5555/3504035.3504406}, which first obtain reliable negative samples from the unlabeled data and then train a binary classifier; however, the performance of these two-step heuristic classifiers is limited by whether the selected samples are correct or not.
Besides the two-step methods, this weakly supervised task can be tackled by one-step methods, by cost-sensitive based methods~\cite{9201373,5559411,LU2021112584}, label disambiguation based methods~\cite{ijcai2019-590}, and density ratio estimation-based methods~\cite{kato2018learning}.
Furthermore, the methods based on risk estimation are some of the most theoretically and practically effective methods~\cite{nnpu,ZHAO2022328,LI2022102947,honecls,puet,distpu}.
The imbalanced PU learning has attracted attention recently~\cite{ijcai2021p412,CHEN2021229}.
Specifically, OC loss~\cite{honecls} has been proposed to solve the imbalance problem in HSI.
However, most of these methods assume that the true $\pi_p$ is available in advance, which is difficult to estimate from HSI with inter-class similarity and intra-class variation.

Learning from PU data without a class prior has recently received attention~\cite{PAN,vPU,p3mix}.
A convex formulation was proposed in~\cite{10.5555/3491440.3491719}.
However, this was based on unbiased risk estimation, and conflicted with the flexible neural networks~\cite{nnpu}.
Predictive adversarial networks (PAN) transform the generator in the generative adversarial network into a classifier~\cite{PAN} to learn from PU data.
A heuristic mixup technique is proposed in~\cite{p3mix}.
The vPU~\cite{vPU} is based on the variational principle.
However, the performance of these methods is unsatisfactory with limited labeled samples, and the problem of unlabeled data dominating the optimization process still exists with vPU.

\paragraph{Other Weakly Supervised Learning Methods}
Label noise representation learning and semi-supervised learning are related to this paper.

The problem of PU learning can be regarded as label noise representation learning, if the unlabeled samples are regarded as noisy negative data.
The adverse effects of noisy labels can be mitigated in three directions: data, optimization policy, and objective~\cite{2020arXiv201104406H}.
For the data, the insight is to link the noisy class posterior and clean class posterior by a noise transition matrix~\cite{c51d68a3106242f08ed001d0c46320b3,goldberger2017training,pmlr-v119-lukasik20a}.
However, the underlying noise transfer pattern is also difficult to estimate.
The dynamic optimization process of the deep neural networks is the key to the optimization policy, such as self-training~\cite{pmlr-v80-jiang18c} and co-training~\cite{NEURIPS2018_a19744e2,pmlr-v97-yu19b}.
However, the noise rate is difficult to estimate.
Mitigating noisy labels from the objective function is consistent with the purpose of this paper, and some loss functions that are robust to noisy labels have in fact been proposed~\cite{10.5555/3298483.3298518,GCE,SCE,TCE}.

The problem of semi-supervised learning is to learn from labeled and unlabeled data~\cite{NEURIPS2020_06964dce,9879201}, in the context of binary classification, the labeled data contains positive and negative data.
However, PU learning is a more challenging task due to the lack of negative samples.
\section{Class Prior-Free PU Learning Framework with Taylor Variational Loss}\label{sec:algorithm}
\begin{figure}[!t]
  \centering
  \includegraphics[width=0.9\columnwidth]{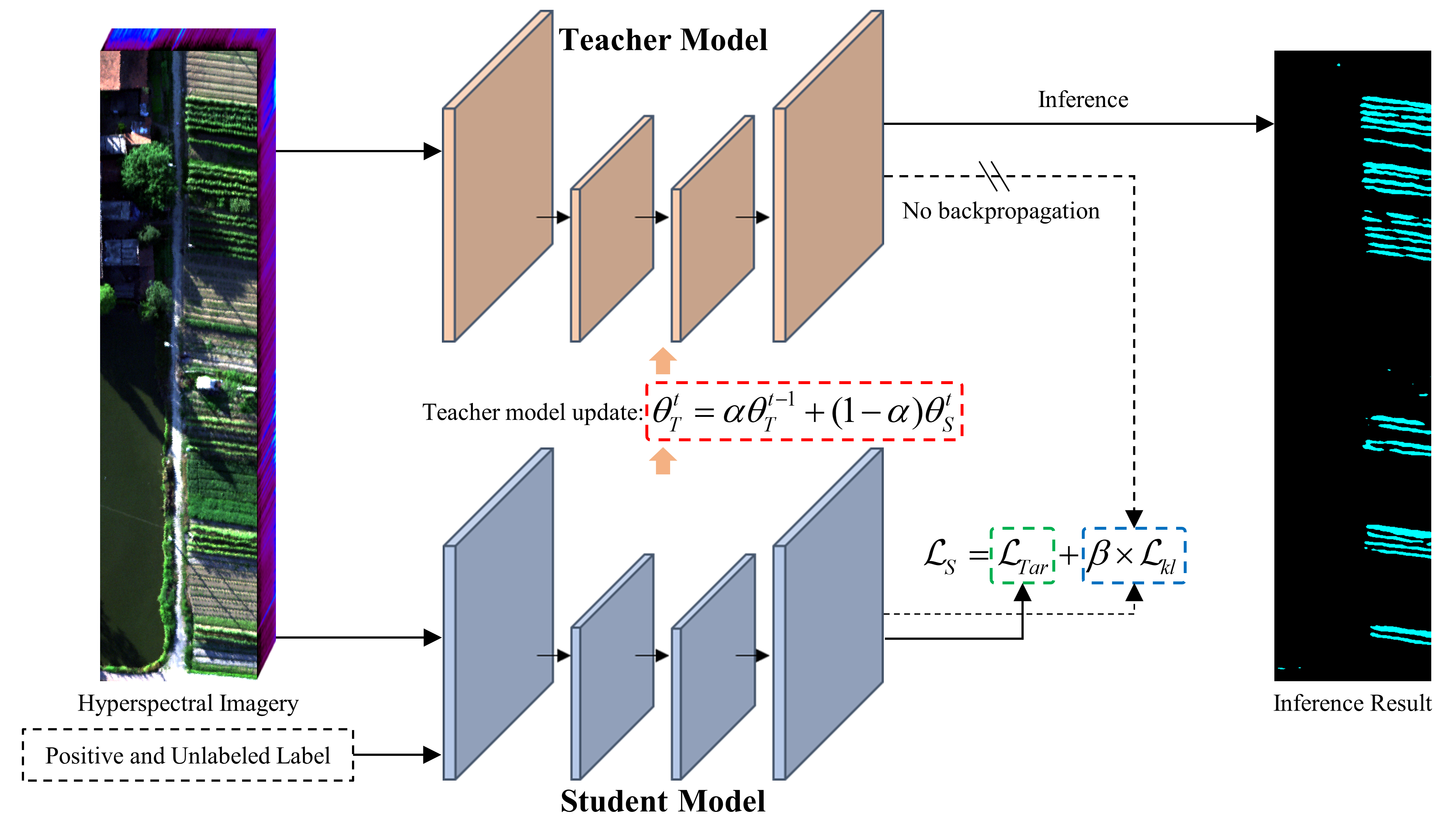}
  \caption{\emph{T-HOneCls}: A Taylor series expansion-based variational framework for HSI PU learning.}
  \label{fig:framework}
\end{figure}

The proposed PU learning framework (dubbed \emph{T-HOneCls}) is described in this section~(Fig.~\ref{fig:framework}).
The proposed \emph{Taylor variational loss} is responsible for the task of learning from PU data without a class prior.
The \emph{self-calibrated optimization} is proposed to stabilize the training process by taking advantage of the supervisory signals from the network itself with a large pool of unlabeled data.

\subsection{Taylor Variational Loss}

\paragraph{Preliminaries}
The spaces of the input and the output are denoted as $X \in R^d$ and $Y \in \left\{+1,-1\right\}$, respectively.
The joint density of $(X,Y)$ is $p(x,y)$.
The marginal distributions of the positive, negative, and unlabeled classes are recorded as $P_{p}(x)=P(x|y=+1)$, $P_{n}(x)=P(x|y=-1)$, and $P(x)$, respectively.
Let $\mathcal{P}=\left\{x_i\right\}^{N_p}_{i=1}\stackrel{\text{i.i.d}}{\sim}P_p(x)$ and $\mathcal{U}=\left\{x_{i}\right\}^{N_u}_{i=1}\stackrel{\text{i.i.d}}{\sim}P(x)$ are the positive and unlabeled dataset, respectively.
For simplicity, $f(x; \theta)$ is denoted as $f(x)$, where $\theta$ represents the parameters of the neural network. 
The PU classifier aims to obtain a parametric classifier, i.e., $f(x)$, from the Bayesian classifier, i.e., $f^*(x)=P(y=+1|x)$, from $\mathcal{P}$ and $\mathcal{U}$.

The estimated positive distribution, i.e., $\hat{P_p}(x)$, can be obtained from the Bayes rule:
\begin{equation}
\begin{aligned}
P_p(x)=\frac{P(y=+1|x)P(x)}{\int P(y=+1|x)P(x)dx} \approx \frac{f(x)P(x)}{E_u[f(x)]} \triangleq \hat{P_p}(x).
\label{eq:estimated_positive_distribution}
\end{aligned}
\end{equation}
If a set $\mathcal{A}$ exists and it satisfies the condition of $\int_{\mathcal{A}}P_p(x)dx>0$ and $f^*(x)=1, {\forall} x {\in}\mathcal{A}$, $P_p(x)=\hat{P_p}(x)$ if and only if $f(x)=f^*(x)$~\cite{vPU}.
The Kullback-Leibler (KL) divergence can be used to estimate the approximate quality of $\hat{P_p}(x)$, and the variational approach can be described as follows:
\begin{equation}
KL(P_p(x)||\hat{P_p}(x))=\mathcal{L}_{var}(f(x))-\mathcal{L}_{var}(f^*(x)),
\label{eq:kl_in_var}
\end{equation}
where
\begin{equation}
\mathcal{L}_{var}(f(x))=\log(E_u[f(x)])-E_p[\log(f(x))].
\label{eq:variational_loss}
\end{equation}
For completeness of this paper, the proof of Eq.~\ref{eq:kl_in_var} is attached to Appendix 1.

According to the non-negative property of KL divergence, $\mathcal{L}_{var}(f(x))$ is the variational upper bound of $\mathcal{L}_{var}(f^*(x))$, and the minimization of Eq.~\ref{eq:kl_in_var} can be achieved by minimizing Eq.~\ref{eq:variational_loss}, which can be calculated from the empirical averages over $\mathcal{P}$ and $\mathcal{U}$ without a class prior by
\begin{equation}
\hat{\mathcal{L}}_{var}(f(x))=\log({\frac{\sum\limits_{i=1}^{n_u}f(x_i^u)}{n_u}})-\frac{\sum\limits_{i=1}^{n_p}\log(f(x_i^p))}{n_p},
\label{eq:empirical_variational_loss}
\end{equation}
where $n_p$ and $n_u$ are the number of positive and unlabeled samples in a batch, respectively.
In other words, the classifier can be obtained by minimizing Eq.~\ref{eq:empirical_variational_loss}, without $\pi_p$.

\begin{figure*}[!t]
    \centering
    \subfloat[\small{Positive loss of the variational classifier}]{
    \label{fig:positive_vpu_loss}
    \includegraphics[width=0.3\textwidth]{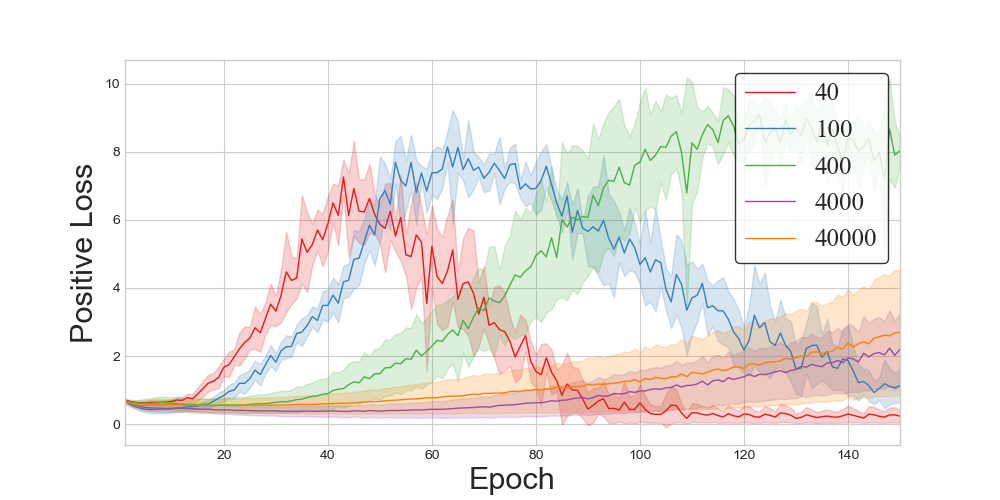}}
    \subfloat[\small{Total loss of the variational classifier}]{
    \label{fig:total_vpu_loss}
    \includegraphics[width=0.3\textwidth]{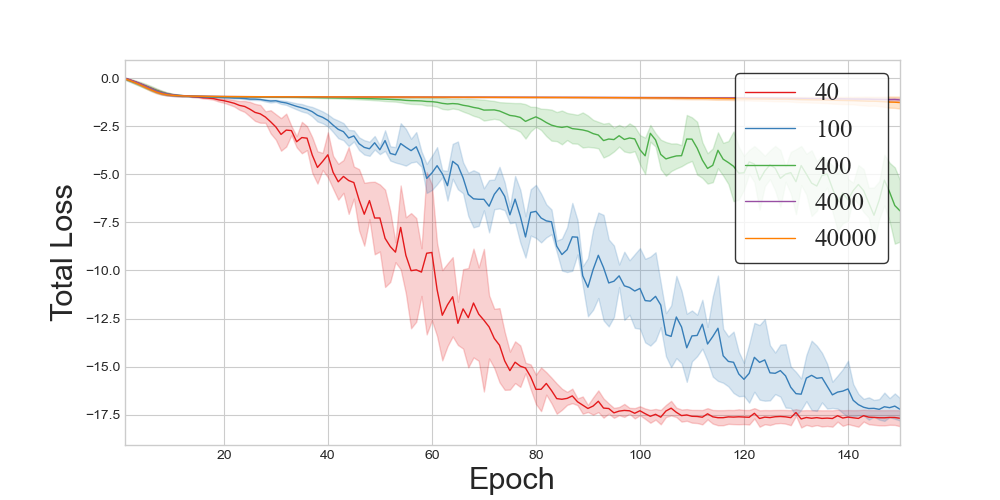}}
    \subfloat[\small{F1-score of the variational classifier}]{
    \label{fig:f1_vpu}
    \includegraphics[width=0.3\textwidth]{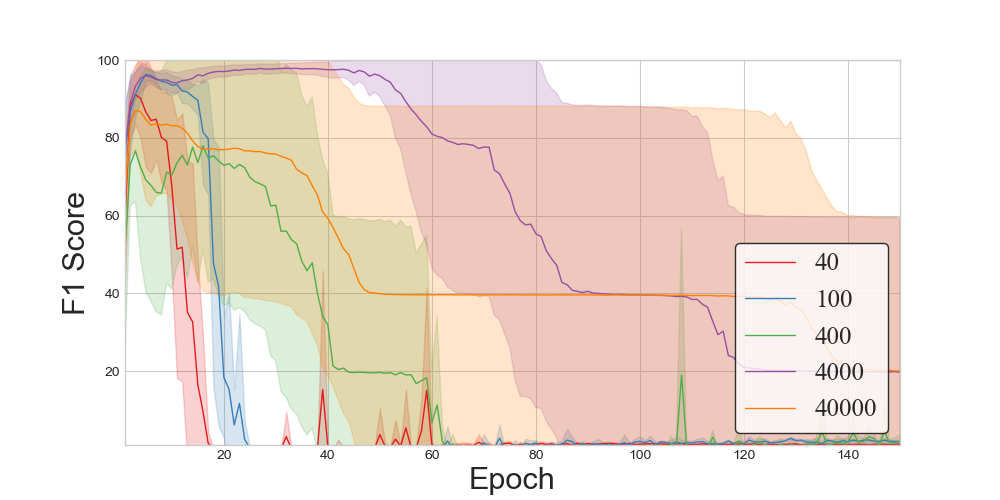}}
    
    \subfloat[\small{Positive loss of \emph{T-HOneCls}}]{
    \label{fig:positive_t_loss}
    \includegraphics[width=0.3\textwidth]{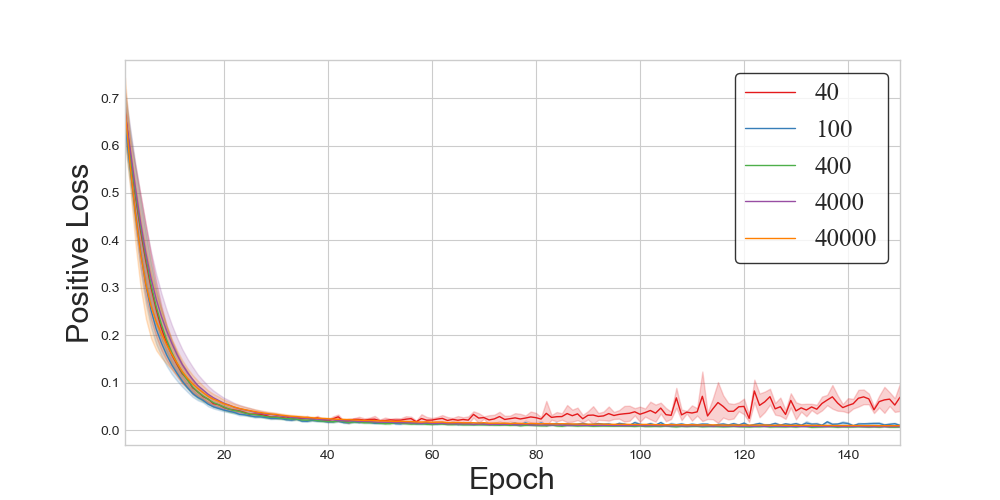}}
    \subfloat[\small{Total loss of \emph{T-HOneCls}}]{
    \label{fig:total_t_loss}
    \includegraphics[width=0.3\textwidth]{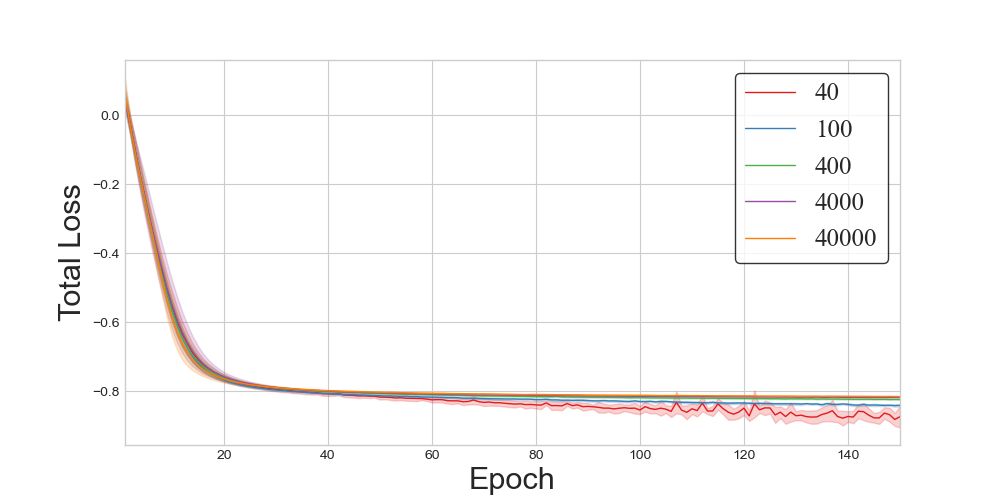}}
    \subfloat[\small{F1-score of \emph{T-HOneCls}}]{
    \label{fig:f1_t}
    \includegraphics[width=0.3\textwidth]{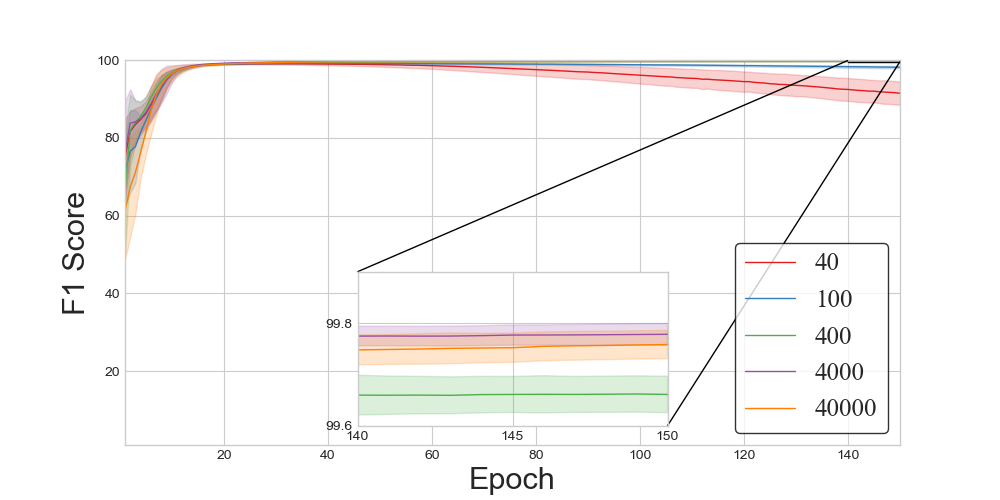}}
    \caption{The curves of loss and F1-score of the variational classifier and \emph{T-HOneCls} with different positive samples in the training stage (taking the cotton in the HongHu dataset as an example).
    The first row show the curves of the variational classifier, and the second row show the curves of the classifier proposed in this paper. The less positive class training data, the faster the variational model collapses.}
    \label{fig:loss_and_f1}
\end{figure*}

\paragraph{Theoretical Analysis of Variational Loss}
The robustness of the variational loss to negative label noise is first analyzed in this subsection, and then a novel insight is proposed to demonstrate that the bottleneck of variational loss is the unlabeled data dominating the training process.

The robustness of variational loss can be obtained by comparing it with cross-entropy loss ($\hat{\mathcal{L}}_{ce}$),
\begin{equation}
\hat{\mathcal{L}}_{ce}(f(x))=-\frac{\sum\limits_{i=1}^{n_n}\log(1-f(x_i^n))}{k}-\frac{\sum\limits_{i=1}^{n_p}\log(f(x_i^p))}{k},
\label{eq:bce_loss}
\end{equation}
where $n_n$ is the number of negative samples in a batch, and $k=n_p+n_n$.

\textit{The first characteristic of variational loss is robustness to negative label noise, which can be analyzed from the weight of the gradient.}
The gradients of the cross-entropy loss and the variational loss are shown in Eq.~\ref{eq:bce_loss_gradient} and Eq.~\ref{eq:variational_loss_gradient}, respectively.
The unlabeled data are treated as noisy negative data in Eq.~\ref{eq:bce_loss_gradient}.
\begin{equation}
    \frac{\partial \hat{\mathcal{L}}_{ce}f(x)}{\partial \theta}={\sum\limits_{i=1}^{n_u}\frac{\nabla_\theta f(x_i^u)}{k(1-f(x_i^u))}}\\ -{\sum\limits_{i=1}^{n_p}\frac{\nabla_\theta f(x_i^p)}{kf(x_i^p)}},
    \label{eq:bce_loss_gradient}
\end{equation}

\begin{equation}
    \frac{\partial \hat{\mathcal{L}}_{var}f(x)}{\partial \theta}={\sum\limits_{i=1}^{n_u}\frac{\nabla_\theta f(x_i^u)}{\sum\limits_{i=1}^{n_u}f(x_i^u)}}-{\sum\limits_{i=1}^{n_p}\frac{\nabla_\theta f(x_i^p)}{n_pf(x_i^p)}}.
    \label{eq:variational_loss_gradient}
\end{equation}
By calculating the gradient of a batch of data from Eq.~\ref{eq:bce_loss_gradient}, the positive data labeled as unlabeled will be given a larger weight if the classifier correctly identifies the sample, and then the neural network will overfit the sample with the wrong label.
However, the variational loss treats each unlabeled sample fairly by assigning the same weight $1/{\sum\limits_{i=1}^{n_u}f(x_i^u)}$, to each unlabeled sample from Eq.~\ref{eq:variational_loss_gradient}, which can alleviate the classifier from overfitting these mislabeled positive samples.

\textit{The second characteristic of variational loss is the problem of the unlabeled data dominating the optimization process, which makes it difficult for neural networks to find a balance between the underfitting and overfitting of positive data.}
This phenomenon can be demonstrated by studying the dynamic changes of the positive part of the variational loss (dubbed positive loss) (Fig.~\ref{fig:loss_and_f1}).
As shown in Fig.~\ref{fig:total_vpu_loss}, although the total loss ($\hat{\mathcal{L}}_{var}(f(x))$) decreases as the training progresses, the positive loss shows an increasing trend in the early training stage (Fig.~\ref{fig:positive_vpu_loss}).
In other words, the unlabeled data dominate the optimization process.
This phenomenon leads to sub-optimal F1-scores and an erratic training process (Fig.~\ref{fig:f1_vpu}).
The number of iterations is uncertain when the unlabeled data dominate training, which leading to a significantly large standard deviation of F1-score in Fig.~\ref{fig:f1_vpu}.
Although the positive loss will decrease when the number of positive data is small, F1-score will not steadily increase, which indicates that the network has changed from underfitting to overfitting of positive data, rapidly.
The smaller the number of positive training samples, the more obvious the instability in the training process, which can be shown in Fig.~\ref{fig:f1_vpu}.

One of the potential factors for training instability is the large weight given to the gradient of the unlabeled data.
A simple example is illustrated: the flexible neural networks can very easily overfit to the training data, which makes $f(x_i^u)$ keep going to 0, and causes the weight of the gradient of the unlabeled samples to keep increasing.
Based on the above analyses, a new loss function is designed in the following.

\paragraph{Taylor Series Expansion for Variational Loss}
The Taylor series expansion is introduced into the variational principle to reduce the weight of the gradient of the unlabeled data and simultaneously satisfy the variational principle, that is, the loss should be greater than or equal to the variational upper bound ($\mathcal{L}_{var}$).

If a given $h(x)$ is differentiable at $x=x_0$ to order $o$, the Taylor series of $h(x)$ is:
\begin{equation}
h(x)=\sum\limits_{i=0}^{\infty} \frac{h^{(i)}(x_0)}{i!}(x-x_0)^i,
\end{equation}
where the $i$-th order derivative of $h(x)$ at $x_0$ is $h^{(i)}(x_0)$.
If the $h(x)$ is defined as $h(x)=\log(x)$, then we set $x_0=1$, and for $\forall i \geq 1$,
\begin{equation}
    h^{(i)}(x_0=1)=(-1)^{i-1}(i-1)!,
\end{equation}
then the $\log(E_u[f(x)])$ can be expressed as
\begin{equation}
    \log(E_u[f(x)])=\sum\limits_{i=1}^{\infty} - \frac{(1-E_u[f(x)])^i}{i}.
    \label{eq:unlabeled_taylor_vpu}
\end{equation}

If the finite terms are reserved, the variational loss can be approximated as
\begin{equation}
\mathcal{L}_{Tar}(f(x))=\sum\limits_{i=1}^{o} - \frac{(1-E_u[f(x)])^i}{i}-E_p[\log(f(x))],
\label{eq:taylor_variational_loss}
\end{equation}
where $o \in \mathcal{N}_+$ denotes the order of the Taylor series.
The \emph{Taylor variational loss} can be calculated from the empirical averages over $\mathcal{P}$ and $\mathcal{U}$ by
\begin{equation}
\hat{\mathcal{L}}_{Tar}(f(x))=\sum\limits_{i=1}^{o} - \frac{\sigma_u^i}{i}-\frac{\sigma_p}{n_p},
\label{eq:empirical_taylor_variational_loss}
\end{equation}
where $\sigma_u=1-\frac{1}{n_u}\sum\limits_{i=1}^{n_u}f(x_i^u)$ and $\sigma_p=\sum\limits_{i=1}^{n_p}\log(f(x_i^p))$.

The proposed \emph{Taylor variational loss} can effectively alleviate the problem of training instability.
It is obvious that
\begin{equation}
  \mathcal{L}_{Tar}(f(x)) \geq \mathcal{L}_{var}(f(x)).
  \end{equation}
The effectiveness of the \emph{Taylor variational loss} can be further illustrated from the weight of the gradient of the unlabeled data.
The detailed proof is as follows:

If we let
\begin{equation}
\hat{\mathcal{L}}_{Tar-u}(f(x))=\sum\limits_{i=1}^{o} - \frac{\sigma_u^i}{i},
\label{eq:empirical_taylor_variational_loss_unlabeled}
\end{equation}
and then,
\begin{equation}
\frac{\partial \hat{\mathcal{L}}_{Tar-u}f(x)}{\partial \theta}=\frac{1}{n_u}\sum\limits_{i=1}^o\sigma_u^{i-1}\sum\limits_{i=1}^{n_u}\nabla_\theta f(x_i^u).
\label{eq:taylor_variational_loss_unlabeled_gradient_1}
\end{equation}
Given that ${0}\textless{\sum\limits_{i=1}^{n_u}f(x_i^u)}\textless{n_u}$, then
\begin{equation}
\frac{\partial \hat{\mathcal{L}}_{Tar-u}f(x)}{\partial \theta}=\frac{1-\sigma_u^o}{\sum\limits_{i=1}^{n_u}f(x_i^u)}\sum\limits_{i=1}^{n_u}\nabla_\theta f(x_i^u).
\label{eq:taylor_variational_loss_unlabeled_gradient_2}
\end{equation}
More proof of Eq.~\ref{eq:taylor_variational_loss_unlabeled_gradient_2} can be found in Appendix 2.

According to Eq.~\ref{eq:taylor_variational_loss_unlabeled_gradient_2}, as with the variational loss, the \emph{Taylor variational loss} also assigns the same weight to each unlabeled sample, but the weight of the unlabeled sample in the \emph{Taylor variational loss} is less than that in variational loss if the finite terms are reserved, as shown in Eq.~\ref{weight_sub}, which prevents the gradients of the unlabeled samples from being given too much weight and then avoids the unlabeled samples dominating the optimization process of the neural network.
\begin{equation}
\frac{1}{\sum\limits_{i=1}^{n_u}f(x_i^u)}-\frac{\sigma_u^o}{\sum\limits_{i=1}^{n_u}f(x_i^u)} \textless \frac{1}{\sum\limits_{i=1}^{n_u}f(x_i^u)}.
\label{weight_sub}
\end{equation}
As $o$ gets larger, the weight of the gradient of the unlabeled samples in \emph{Taylor variational loss} is convergent to that of variational loss for a given classifier.

\subsection{Self-calibrated Optimization}

\emph{Self-calibrated optimization} is aimed at improving the performance of the classifier from the optimization process by using additional supervisory signals from the neural network itself.
Specifically, \emph{KL-Teacher} is proposed to utilize the memorization ability of the neural network, to stabilize the training process and alleviate the overfitting problem with a large pool of unlabeled data.

The memorization ability~\cite{10.5555/3305381.3305406} of the neural network can also be observed when using variational-based loss to train the neural network.
As the number of training epochs increases, the F1-score of the test set will first rise and then decrease until convergence, as shown by the curves of the F1-score in Fig.~\ref{fig:f1_vpu}, especially when the number of labeled samples is limited (40 labeled samples).

In order to capture the supervisory signal brought by the memorization ability of the neural network, two neural networks with the same architecture are used, with one being the teacher network ($T$) and the other the student network ($S$).
The weights of the teacher network ($\theta_T^t$, where $t$ is the number of iterations) are updated by the exponential moving average (EMA) of the student network, as follows:
\begin{equation}
\theta_T^t = \alpha\theta_T^{t-1}+(1-\alpha)\theta_S^t.
\label{eq:kl_teacher_model_update}
\end{equation}
Due to the utilization of the EMA, the teacher network acts as an ``F1-score filter" and can obtain more stable classification results, which is demonstrated in Section~\ref{sec:results}.

A consistency loss ($\mathcal{L}_{kl}$) based on KL divergence is used to force the teacher network and the student network to have the same output, which can be used as an additional supervisory signal to alleviate the overfitting problem of the student network from a large pool of unlabeled data:
\begin{equation}
\mathcal{L}_{kl} = KL(p_T||p_S)+KL(p_S||p_T),
\label{eq:kl_teacher_loss}
\end{equation}
where $p_T$ and $p_S$ are the probabilistic outputs of the teacher network and the student network, respectively.
The objective function of the student network is:
\begin{equation}
    \mathcal{L}_S=\mathcal{L}_{Tar}+\beta\mathcal{L}_{kl}.
\end{equation}
The output of the teacher network is used as the final classification result.

A detailed description of the training of \emph{T-HOneCls} is provided in Appendix 3.
More ablation experiments about EMA and $\mathcal{L}_{kl}$ can be found in Section~\ref{sec:results}.
\section{Experimental Results and Analysis}\label{sec:results}
\subsection{Experimental Settings}
\paragraph{Datasets}

7 challenging datasets were used, including 3 UAV hyperspectral datasets (HongHu, LongKou, and HanChuan, 15 tasks in total)~\cite{ZHONG2020112012}, 2 HSI classification datasets (Indian Pines and Pavia University, 4 tasks in total) and 2 RGB datasets (CIFAR-10 and STL-10).
More detailed information can be found in Appendix 4.

PU learning on UAV hyperspectral datasets is a challenging task.
These UAV datasets mainly contain visually indistinct crops, and have strong inter-class similarity and intra-class variation.
The UAV HSI along with the ground truth and spectral curves as an example are shown in Appendix 4.
It can be seen that the spectral curves of the vegetation are very similar.
In particular, there are shadows in the HanChuan dataset, which significantly increase the intra-class variability.
In UAV datasets, some ground objects with very high textural and spectral similarity were selected for classification.
For 5 HSI datasets, only 100 positive samples for each class were used to simulate the situation of limited training samples to train the neural network.

CIFAR-10 and STL-10 were used to verify the effectiveness of the proposed $\mathcal{L}_{Tar}$ compared with other state-of-the-art PU learning methods.

\begin{table*}[!hbt]
  \centering
  \resizebox{.98\textwidth}{!}{
    \begin{tabular}{cccccccccc}
      \toprule
      \multicolumn{1}{c|}{\multirow{2}{*}{Class}} & \multicolumn{2}{c|}{Class prior-based classifiers} & \multicolumn{4}{c|}{Label noise representation learning}                & \multicolumn{3}{c}{Class prior-free classifiers}  \\
      \multicolumn{1}{c|}{}                       & nnPU~\cite{nnpu}             & \multicolumn{1}{c|}{OC Loss~\cite{honecls}}    & MSE Loss~\cite{10.5555/3298483.3298518}    & GCE Loss~\cite{GCE}    & SCE Loss~\cite{SCE}    & \multicolumn{1}{c|}{TCE Loss~\cite{TCE}} & PAN~\cite{PAN}          & vPU~\cite{vPU}         & T-HOneCls            \\ \midrule
      Cotton                                      & 99.44(0.32)      & \textbf{99.44(0.25)}            & 17.08(8.25) & 18.39(4.80) & 96.34(2.36) & 20.11(6.31)                   & 16.66(1.40)  & 1.86(0.48)  & 98.15(0.35)          \\
      Rape                                        & 82.06(0.71)      & 81.81(1.23)                     & 96.32(0.72) & 96.69(0.72) & 97.35(0.18) & 97.64(0.12)                   & 77.89(10.17) & 8.31(1.10)  & \textbf{97.81(0.16)} \\
      Chinese cabbage                             & 0.00(0.00)       & 88.06(2.89)                     & 93.61(0.55) & 94.06(0.60) & 93.78(0.63) & 94.19(0.43)                   & 92.31(1.34)  & 24.89(1.22) & \textbf{94.25(0.70)} \\
      Cabbage                                     & 54.20(49.50)     & 89.79(1.27)                     & 99.20(0.21) & 99.10(0.18) & 99.12(0.20) & 99.30(0.08)                   & 98.18(0.28)  & 34.84(2.51) & \textbf{99.37(0.07)} \\
      Tuber mustard                               & 23.99(0.21)      & 23.57(0.22)                     & 95.23(0.66) & 96.05(0.56) & 95.50(0.87) & 96.60(0.11)                   & 92.17(1.79)  & 23.28(1.19) & \textbf{97.38(0.35)} \\ \midrule
      Macro F1                                    & 51.94            & 76.53                           & 80.29       & 80.86       & 96.42       & 81.57                         & 75.44        & 18.64       & \textbf{97.39}       \\ \midrule
      \multicolumn{3}{l}{Macro F1 of supervised binary classifier}                                     & \multicolumn{7}{c}{75.62}                                                                                                   \\ \bottomrule
      \end{tabular}}
      \caption{The F1-scores for the HongHu dataset}
      \label{tab:HongHu_f1}
\end{table*}

\begin{table*}[!hbt]
  \centering
  \resizebox{.98\textwidth}{!}{
    \begin{tabular}{cccccccccc}
      \toprule
      \multicolumn{1}{c|}{\multirow{2}{*}{Class}} & \multicolumn{2}{c|}{Class prior-based classifiers} & \multicolumn{4}{c|}{Label noise representation learning}                         & \multicolumn{3}{c}{Class prior-free classifiers}  \\
      \multicolumn{1}{c|}{}                       & nnPU~\cite{nnpu}            & \multicolumn{1}{c|}{OC Loss~\cite{honecls}}     & MSE Loss~\cite{10.5555/3298483.3298518}             & GCE Loss~\cite{GCE}    & SCE Loss~\cite{SCE}    & \multicolumn{1}{c|}{TCE Loss~\cite{TCE}} & PAN~\cite{PAN}          & vPU~\cite{vPU}         & T-HOneCls            \\ \midrule
      Strawberry                                  & 89.16(1.49)     & 89.52(1.54)                      & 33.69(5.71)          & 34.56(2.53) & 92.44(0.96) & 77.69(18.03)                  & 30.95(0.88)  & 9.40(0.97)  & \textbf{94.58(1.28)} \\
      Cowpea                                      & 59.66(3.63)     & 58.97(3.56)                      & 46.55(3.39)          & 46.27(2.38) & 70.98(7.69) & 56.82(3.09)                   & 43.95(1.08)  & 12.83(1.00) & \textbf{90.31(1.13)} \\
      Soybean                                     & 43.63(3.14)     & 42.34(1.06)                      & 97.42(0.94)          & 97.26(1.06) & 97.19(1.11) & 98.55(0.59)                   & 86.74(4.51)  & 38.73(2.36) & \textbf{99.13(0.28)} \\
      Watermelon                                  & 11.76(0.36)     & 12.23(0.46)                      & \textbf{94.02(0.74)} & 93.79(0.98) & 93.45(0.94) & 92.67(0.84)                   & 91.99(0.45)  & 54.77(2.43) & 92.99(0.90)          \\
      Road                                        & 0.00(0.00)      & 89.40(4.34)                      & 76.54(4.98)          & 74.53(3.88) & 85.71(1.84) & 86.29(2.06)                   & 61.56(1.93)  & 25.02(1.63) & \textbf{91.73(1.06)} \\
      Water                                       & 95.25(0.81)     & 94.90(0.63)                      & 87.52(9.20)          & 92.12(5.26) & 96.97(0.49) & 94.15(4.70)                   & 73.08(24.40) & 1.43(0.98)  & \textbf{98.37(0.32)} \\ \midrule
      Macro F1                                    & 49.91           & 64.56                            & 72.62                & 73.09       & 89.46       & 84.36                         & 64.71        & 23.70       & \textbf{94.52}       \\ \midrule
      \multicolumn{3}{l}{Macro F1 of supervised binary classifier}                                     & \multicolumn{7}{c}{66.96}                                                                                                            \\ \bottomrule
      \end{tabular}}
    \caption{The F1-scores for the HanChuan dataset}
    \label{tab:HanChuan_f1}
\end{table*}

\paragraph{Training Details}
As for hyperspectral datasets, following~\cite{honecls}, this paper used FreeOCNet as the fully convolutional neural network.
As shown in~Appendix 5, FreeOCNet includes an encoder, decoder, and lateral connection.
More details about FreeOCNet can be found in~\cite{honecls}.
In order to make a fair comparison, all the methods used the same network and the same common hyperparameters.
If not specified, the order of the Taylor expansion in \emph{T-HOneCls} is 2, and $\alpha=0.99$.
$\beta=0.5$ in the HongHu, LongKou, Indian Pines and Pavia University datasets, and $\beta=0.2$ in the HanChuan dataset.
As for RGB datasets, 7-layer CNN was used for CIFAR-10 and STL-10.
The settings of these common hyperparameters are listed in Appendix 4.
The experiments were conducted using an NVIDIA RTX 3090 GPU.

\begin{table*}[!hbt]
  \centering
  \resizebox{.98\textwidth}{!}{
    \begin{tabular}{cccccccccc}
      \toprule
      \multicolumn{1}{c|}{\multirow{2}{*}{Class}} & \multicolumn{2}{c|}{Class prior-based classifiers} & \multicolumn{4}{c|}{Label noise representation learning}                & \multicolumn{3}{c}{Class prior-free classifiers}  \\
      \multicolumn{1}{c|}{}                       & nnPU~\cite{nnpu}             & \multicolumn{1}{c|}{OC Loss~\cite{honecls}}    & MSE Loss~\cite{10.5555/3298483.3298518}    & GCE Loss~\cite{GCE}    & SCE Loss~\cite{SCE}    & \multicolumn{1}{c|}{TCE Loss~\cite{TCE}} & PAN~\cite{PAN}         & vPU~\cite{vPU}          & T-HOneCls            \\ \midrule
      Corn                                        & 98.54(2.24)      & 99.67(0.11)                     & 99.44(0.27) & 99.16(0.25) & 98.50(0.87) & 98.82(0.70)                   & 97.16(2.10) & 8.54(1.03)   & \textbf{99.70(0.12)} \\
      Sesame                                      & 10.97(24.52)     & 75.95(2.78)                     & 99.77(0.07) & 99.77(0.09) & 99.78(0.03) & 99.79(0.09)                   & 99.73(0.04) & 67.99(13.73) & \textbf{99.82(0.07)} \\
      Broad-leaf soybean                          & 84.69(1.11)      & 88.02(0.26)                     & 81.98(2.84) & 87.29(1.67) & 87.03(3.36) & 74.94(3.48)                   & 58.23(6.90) & 4.47(0.25)   & \textbf{92.64(0.89)} \\
      Rice                                  & 0.00(0.00)       & \textbf{99.70(0.39)}            & 98.94(0.24) & 99.19(0.14) & 99.16(0.24) & 98.78(0.84)                   & 98.63(0.40) & 34.94(1.28)  & 99.50(0.16)          \\ \midrule
      Macro F1                                    & 48.55            & 90.84                           & 95.03       & 96.35       & 96.12       & 93.09                         & 88.44       & 28.98        & \textbf{97.92}       \\ \midrule
      \multicolumn{3}{l}{Macro F1 of supervised binary classifier}                                     & \multicolumn{7}{c}{90.49}                                                                                                   \\ \bottomrule
      \end{tabular}}
      \caption{The F1-scores for the LongKou dataset}
      \label{tab:LongKou_f1}
\end{table*}

\begin{table*}[!hbt]
  \centering
  \resizebox{.98\textwidth}{!}{
    \begin{tabular}{cccccccccc}
      \toprule
      \multicolumn{1}{c|}{\multirow{2}{*}{Class}} & \multicolumn{2}{c|}{Class prior-based classifiers} & \multicolumn{4}{c|}{Label noise representation learning}                & \multicolumn{3}{c}{Class prior-free classifiers} \\
      \multicolumn{1}{c|}{}                       & nnPU~\cite{nnpu}            & \multicolumn{1}{c|}{OC Loss~\cite{honecls}}     & MSE Loss~\cite{10.5555/3298483.3298518}    & GCE Loss~\cite{GCE}    & SCE Loss~\cite{SCE}    & \multicolumn{1}{c|}{TCE Loss~\cite{TCE}} & PAN~\cite{PAN}         & vPU~\cite{PAN}         & T-HOneCls            \\ \midrule
      India Pines-2                                & 42.30(0.73)     & 43.14(0.96)                      & 85.30(1.19) & 86.16(2.19) & 86.89(0.77) & 88.60(1.45)                   & 82.54(1.45) & 8.44(1.72)  & \textbf{93.40(0.50)} \\
      India Pines-11                               & 63.35(1.01)     & 62.88(0.46)                      & 75.95(2.64) & 77.04(2.30) & 83.65(1.34) & 83.03(1.73)                   & 65.22(3.69) & 3.40(0.62)  & \textbf{91.86(1.14)} \\
      Pavia University-2                                    & 89.17(2.60)     & 90.75(0.80)                      & 93.52(1.24) & 91.29(1.45) & 92.38(2.54) & 90.41(1.14)                   & 89.92(3.49) & 10.74(2.32) & \textbf{95.01(1.04)} \\
      Pavia University-8                                    & 0.00(0.00)      & 82.63(3.46)                      & 90.90(0.67) & 91.27(1.46) & 88.67(1.46) & \textbf{92.05(0.77)}          & 87.08(1.59) & 37.46(2.20) & 91.89(1.81)          \\ \bottomrule
      \end{tabular}}
    \caption{The F1-scores for the Indian Pines and Pavia University datasets}
    \label{tab:ip_pu_f1}
\end{table*}

\paragraph{Metrics}
The F1-score were selected as the metric to measure the performance in HSI datasets.
The precision and recall are shown in Appendix 6 as supplements.
The macro F1-score is the average of the F1-scores over the selected classes, which can measure the robustness of a classifier on different ground objects.
The overall accuracy (OA) were selected as the metric in RGB datasets.
Without special instructions, all the experiments were repeated five times, and the mean and standard deviation values are reported.

\paragraph{Methods}
There were three types of comparison algorithms in HSI datasets.
Firstly, the proposed method---\emph{T-HOneCls}---is compared with the class prior based classifiers, i.e., nnPU~\cite{nnpu} and OC Loss~\cite{honecls}.
The class priors were estimated by the KMPE~\cite{kmpe}.
Methods of label noise representation learning were also compared, i.e., MSE Loss~\cite{10.5555/3298483.3298518}, GCE Loss~\cite{GCE}, SCE Loss~\cite{SCE}, TCE Loss~\cite{TCE}.
What is more, the proposed method was also compared with the state-of-the-art class prior-free PU classifiers from the machine learning community, i.e., PAN~\cite{PAN} and vPU~\cite{vPU}.
As a supplement, unlabeled data is used as negative class to illustrate that the performance of supervised binary classifier is limited in one-class scenarios.

As for RGB datasets, the proposed $\mathcal{L}_{Tar}$ is compared with other state-of-the-art PU learning methods: nnPU~\cite{nnpu}, PUET~\cite{puet}, DistPU~\cite{distpu}, P3MIX~\cite{p3mix} and $\mathcal{L}_{var}$~\cite{vPU}.

\subsection{Results on Hyperspectral Datasets}

The results of hyperspectral data are listed in Table~\ref{tab:HongHu_f1}-Table~\ref{tab:ip_pu_f1}.
Limited by the number of pages, the distribution maps are shown in Appendix 6.

From the macro F1-score, \emph{T-HOneCls} achieves the best results in all UAV datasets, which fully demonstrates the robustness of the proposed algorithm.
A more detailed analysis follows: 1) It is clear that, without the limitation of the class prior, the macro F1-score of \emph{T-OneCls} is significantly higher than that of the class prior-based methods.
The class prior estimation for cotton is accurate, and the best F1-score for the cotton is obtained by the class prior-based methods; however, the F1-score drops when the estimated class prior is inaccurate (e.g., tuber mustard).
2) Compared with the label noise representation learning methods, \emph{T-HOneCls} achieves a better F1-score in 17 of the 19 tasks, which indicates the necessity for developing a PU algorithm instead of directly applying the label noise representation learning methods to HSI.
3) Compared with the rencent class prior-free methods proposed by the machine learning community, \emph{T-HOneCls} obtains a better F1-score on all tasks.

Another conclusion is that the proposed \textit{T-HOneCls} can balance the precision and recall.
As shown in Appendix 6, most other methods cannot obtain high precision and recall at the same time, that is, these methods cannot find a balance between the overfitting and underfitting of the training data.
This balance was found by \textit{T-HOneCls}, and a good F1-score was obtained by \textit{T-HOneCls} in all tasks.

\begin{figure}[!t]
  \centering
  \subfloat[\small{CIFAR-10 loss}]{
  \label{fig:loss_cifar10}
  \includegraphics[width=0.45\columnwidth]{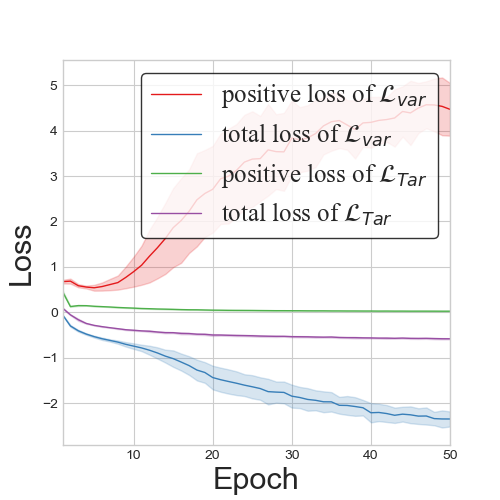}}
  \subfloat[\small{STL-10 loss}]{
  \label{fig:loss_stl10}
  \includegraphics[width=0.45\columnwidth]{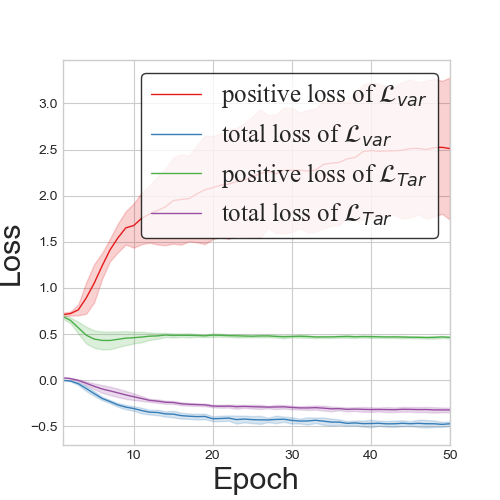}}
  \vspace{-0.4cm}
  \subfloat[\small{CIFAR-10 OA}]{
  \label{fig:oa_cifar10}
  \includegraphics[width=0.45\columnwidth]{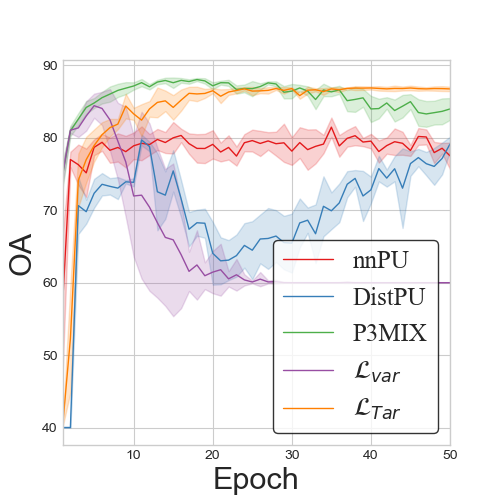}}
  \subfloat[\small{STL-10 OA}]{
  \label{fig:oa_stl10}
  \includegraphics[width=0.45\columnwidth]{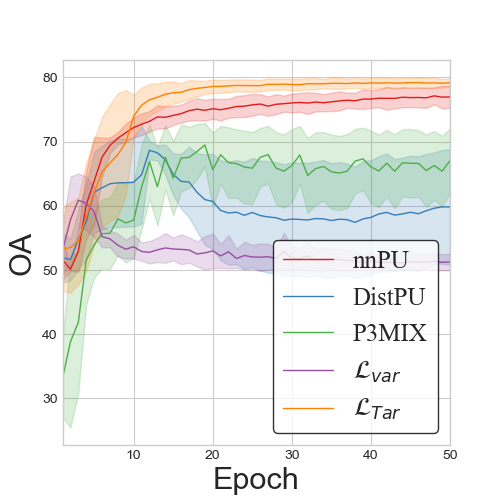}}
  \label{fig:oa_cifar10_stl10}
  \caption{The curves of loss and OA on CIFAR-10 and STL-10 datasets.}
  \label{fig:rgb_loss_and_oa}
\end{figure}

\subsection{Results on CIFAR-10 and STL-10}
The experimental results on RGB datasets show that $\mathcal{L}_{Tar}$ is not limited to hyperspectral data, and $\mathcal{L}_{Tar}$ also performs well in other PU learning tasks.
The OA of $\mathcal{L}_{Tar}$ is better than that of other state-of-the-art PU learning methods (Table~\ref{tab:rgb_f1}), and the curves of loss and OA can also prove the effectiveness of the proposed $\mathcal{L}_{Tar}$ (Fig.~\ref{fig:rgb_loss_and_oa}).

\subsection{Ablation Experiments Analysis}

\paragraph{Analysis of the Training Process and Training Samples}
The curves of \emph{T-HOneCls} for the positive class and the total loss of the different positive training samples of cotton in the HongHu dataset are shown in Fig.~\ref{fig:positive_t_loss} and Fig.~\ref{fig:total_t_loss}, respectively.
The curves of the F1-score are also shown (Fig.~\ref{fig:f1_t}).
\begin{table}[!t]
  \centering
  \resizebox{.98\columnwidth}{!}{  
    \begin{tabular}{cccccccc}
      \toprule
      \multicolumn{1}{c}{Datasets} & nnPU~\cite{nnpu}  & PUET~\cite{puet} & DistPU~\cite{distpu} & P3MIX~\cite{p3mix}       & $\mathcal{L}_{var}$~\cite{vPU} & $\mathcal{L}_{Tar}$ \\ \midrule
      CIFAR-10                      & 77.53(2.04)    & 75.60(0.10)        & 79.15(1.12)      & 83.99(1.68) & 60.00(0.00)         & \textbf{86.76(0.35)}         \\
      STL-10                        & 76.98(1.91)    & 75.67(0.22)         & 59.83(10.03)      & 67.05(5.58) & 51.26(1.46)         & \textbf{79.17(0.71)}         \\ \bottomrule
      \end{tabular}}
      \caption{The OA of different methods on CIFAR-10 and STL-10 datasets.
      Definitions of classes (`Positive' vs `Negative') are as follows: CIFAR-10: `0,1,8,9' vs `2,3,4,5,6,7'. STL-10: `0,2,3,8,9' vs `1,4,5,6,7'.}
      \label{tab:rgb_f1}
  \end{table}
The variational loss using fewer training samples will lead to the gradient domination optimization process of unlabeled samples at the beginning of the training, which makes the loss of positive class rise at the beginning of the training.
Although the loss of the positive samples decreases as the training progresses, for example, 40, 100, or 400, the F1-score is unstable, and determining the optimal training epoch is very challenging without using additional data.
The total loss of cotton of vPU shows large reduction in Fig.~\ref{fig:loss_and_f1}, however, the F1 (1.86) is very poor, which is because vPU overfits the noisy negative data (i.e., unlabeled data).
These shortcomings can be solved by the proposed $\mathcal{L}_{Tar}$ due to the reduction of the weight of the gradient of unlabeled data.
More analysis can be found in Appendix 7.

\begin{figure}[!t]
  \centering
  \includegraphics[width=0.98\columnwidth]{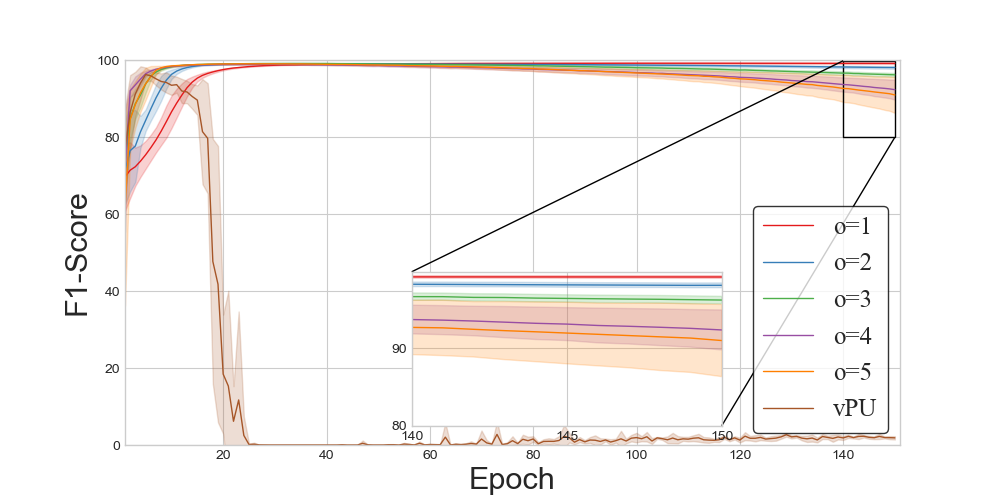}
  \caption{The F1-score curves (cotton in the HongHu dataset) for the different order of the Taylor series.}
  \label{fig:order_experiments}
\end{figure}

\paragraph{Analysis of the Order of the Taylor Series}
One of the contributions of this paper is that we point out that the reason for the poor performance of variational loss is that the gradient of the unlabeled data is given too much weight, which can be tackled by the proposed \emph{Taylor variational loss}.
The order of the Taylor expansion is analyzed as a hyperparameter in this subsection, and the F1-score curves of cotton in the HongHu dataset are shown in Fig.~\ref{fig:order_experiments} as an example.
Five other ground objects were also analyzed, and the results are displayed in Appendix 8.
As shown in Fig.~\ref{fig:order_experiments}, the neural networks converge to a poor result with variational loss.
An empirical conclusion can be obtained from the order analysis: the higher the order of the Taylor expansion, the faster the neural network converges.
However, the rapid convergence of the neural network can lead to overfitting.
In other words, the classification results will rise first and then decline with the progress of the training.

\paragraph{Analysis of KL-Teacher}
This subsection analyzes the advantages of the proposed self-calibrated optimization.
Three ground objects from the three datasets were selected as examples to demonstrate the advantages of self-calibration optimization.
The F1-score curves of cowpea in the HanChuan dataset are shown in Fig.~\ref{fig:kl_teacher_experiments} and other classes are shown in Appendix 9.

It can be seen from Table~\ref{tab:KL-Teacher_f1} that the training is failed, if $\mathcal{L}_{var}$ with self-calibrated optimization is used.
It can be seen from Fig.~\ref{fig:kl_teacher_experiments} that the F1-score fluctuates greatly when only stochastic gradient descent is used to optimize the \emph{Taylor variational loss}.
The EMA has the function of an ``F1-score filter'', which makes the F1-score of the teacher model more stable.
The EMA allows the teacher model to lag behind the student model, and due to the memorization ability of the neural network, the F1-score of the lagged neural network is better than that of the student network at the later stage of training.
The use of consistency loss can promote the output of the student model to approximate that of the teacher model with a large pool of unlabeled data, so as to alleviate the overfitting problem.
If L2 loss ($\mathcal{L}_2$) is regarded as the consistency loss, it is equivalent to Mean-Teacher~\cite{mean_teacher} being used.
However, according to the results listed in Table~\ref{tab:KL-Teacher_f1}, $\mathcal{L}_{kl}$ can more effectively alleviate the overfitting of the student model.

\begin{figure}[!t]
  \centering
  \includegraphics[width=0.98\columnwidth]{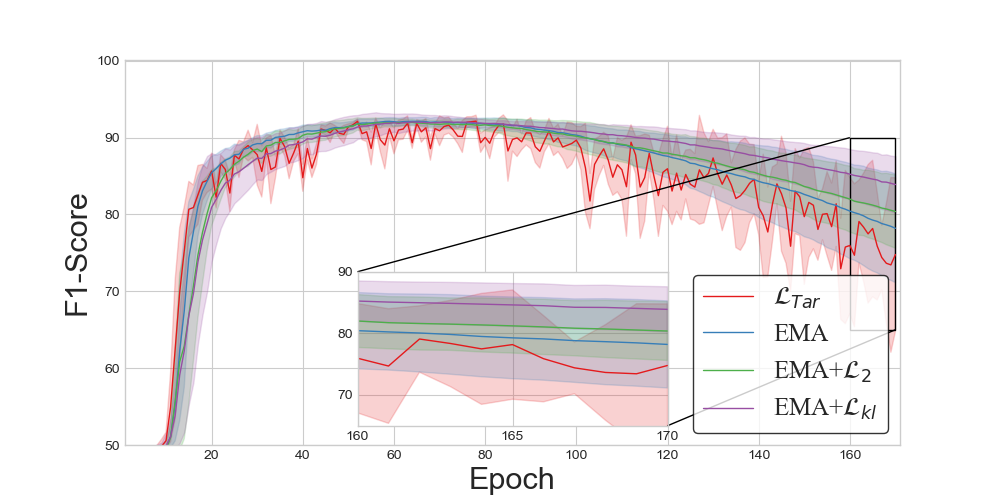}
  \caption{The F1-score curves (cowpea in the HanChuan dataset, o=5) for the different components of \emph{KL-Teacher}.}
  \label{fig:kl_teacher_experiments}
\end{figure}

\begin{table}[!t]
  \centering
  \resizebox{.95\columnwidth}{!}{  
    \begin{tabular}{ccccccc}
      \toprule
      \multirow{2}{*}{Class}                                                                     & \multirow{2}{*}{Order} & \multirow{2}{*}{$\mathcal{L}$} & \multicolumn{3}{c}{Self-calibrated optimization}                                  & \multirow{2}{*}{F1-score} \\ \cmidrule{4-6}
                                                                                                 &                        &                                & EMA                       & $\mathcal{L}_2$           & $\mathcal{L}_{kl}$        &                           \\ \midrule
      \multirow{9}{*}{\begin{tabular}[c]{@{}c@{}}Cotton\end{tabular}}              & -                      & $\mathcal{L}_{var}$            & \checkmark &                           & \checkmark & 0.00                      \\ \cmidrule{2-7} 
                                                                                                 & \multirow{4}{*}{2}     & $\mathcal{L}_{Tar}$            &                           &                           &                           & 97.51                     \\
                                                                                                 &                        & $\mathcal{L}_{Tar}$            & \checkmark &                           &                           & 97.58                     \\
                                                                                                 &                        & $\mathcal{L}_{Tar}$            & \checkmark & \checkmark &                           & 97.61                     \\
                                                                                                 &                        & $\mathcal{L}_{Tar}$            & \checkmark &                           & \checkmark & \textbf{98.15}            \\ \cmidrule{2-7} 
                                                                                                 & \multirow{4}{*}{5}     & $\mathcal{L}_{Tar}$            &                           &                           &                           & 72.01                     \\
                                                                                                 &                        & $\mathcal{L}_{Tar}$            & \checkmark &                           &                           & 84.27                     \\
                                                                                                 &                        & $\mathcal{L}_{Tar}$            & \checkmark & \checkmark &                           & 81.25                     \\
                                                                                                 &                        & $\mathcal{L}_{Tar}$            & \checkmark &                           & \checkmark & \textbf{91.01}            \\ \midrule
      \multirow{9}{*}{\begin{tabular}[c]{@{}c@{}}Broad-leaf soybean\end{tabular}} & -                      & $\mathcal{L}_{var}$            & \checkmark &                           & \checkmark & 0.12                      \\ \cmidrule{2-7} 
                                                                                                 & \multirow{4}{*}{2}     & $\mathcal{L}_{Tar}$            &                           &                           &                           & 90.74                     \\
                                                                                                 &                        & $\mathcal{L}_{Tar}$            & \checkmark &                           &                           & 91.22                     \\
                                                                                                 &                        & $\mathcal{L}_{Tar}$            & \checkmark & \checkmark &                           & 91.42                     \\
                                                                                                 &                        & $\mathcal{L}_{Tar}$            & \checkmark &                           & \checkmark & \textbf{92.64}            \\ \cmidrule{2-7} 
                                                                                                 & \multirow{4}{*}{5}     & $\mathcal{L}_{Tar}$            &                           &                           &                           & 81.06                     \\
                                                                                                 &                        & $\mathcal{L}_{Tar}$            & \checkmark &                           &                           & 81.61                     \\
                                                                                                 &                        & $\mathcal{L}_{Tar}$            & \checkmark & \checkmark &                           & 81.78                     \\
                                                                                                 &                        & $\mathcal{L}_{Tar}$            & \checkmark &                           & \checkmark & \textbf{82.79}            \\ \midrule
      \multirow{9}{*}{\begin{tabular}[c]{@{}c@{}}Cowpea\end{tabular}}            & -                      & $\mathcal{L}_{var}$            & \checkmark &                           & \checkmark & 4.00                      \\ \cmidrule{2-7} 
                                                                                                 & \multirow{4}{*}{2}     & $\mathcal{L}_{Tar}$            &                           &                           &                           & 88.87                     \\
                                                                                                 &                        & $\mathcal{L}_{Tar}$            & \checkmark &                           &                           & 88.59                     \\
                                                                                                 &                        & $\mathcal{L}_{Tar}$            & \checkmark & \checkmark &                           & 88.78                     \\
                                                                                                 &                        & $\mathcal{L}_{Tar}$            & \checkmark &                           & \checkmark & \textbf{90.31}            \\ \cmidrule{2-7} 
                                                                                                 & \multirow{4}{*}{5}     & $\mathcal{L}_{Tar}$            &                           &                           &                           & 74.78                     \\
                                                                                                 &                        & $\mathcal{L}_{Tar}$            & \checkmark &                           &                           & 78.20                     \\
                                                                                                 &                        & $\mathcal{L}_{Tar}$            & \checkmark & \checkmark &                           & 80.38                     \\
                                                                                                 &                        & $\mathcal{L}_{Tar}$            & \checkmark &                           & \checkmark & \textbf{83.90}            \\ \bottomrule
\end{tabular}}
\caption{Analysis of KL-Teacher}
\label{tab:KL-Teacher_f1}
\end{table}
\section{Conclusion}
In this paper, we have focused on tackling the problem of limited labeled HSI PU learning without class-prior.
The proposed \emph{Taylor variational loss} is responsible for the task of learning from limited labeled PU data without a class prior.
The \emph{self-calibrated optimization} proposed in this paper is used to stabilize the training process.
The extensive experiments (7 datasets, 21 tasks in total) demonstrated the superiority of the proposed method.

\noindent \textbf{Acknowledgements:}
This work was supported by National Key Research and Development Program of China under Grant No.2022YFB3903502, National Natural Science Foundation of China under Grant No.42325105, 42071350, 42101327, and LIESMARS Special Research Funding.

{\small
\bibliographystyle{ieee_fullname}
\bibliography{bib}
}

\clearpage
\section*{Appendix}
\subsection*{1. Proof of Eq.~\ref{eq:kl_in_var}}
From the definition of KL divergence, Eq.~\ref{eq:kl_in_var} can be formulated as follows:
\begin{equation}\nonumber
\begin{aligned}
&KL(P_p(x)||\hat{P_p}(x))\\
&=E_p[\log(\frac{P_p(x)}{\hat{P_p}(x)})]\\
&=E_p[\log(f^*(x))]-\log(E_u[f^*(x)])\\
&\quad-E_p[\log(f(x))]+\log(E_u[f(x)])\\
&=\mathcal{L}_{var}(f(x))-\mathcal{L}_{var}(f^*(x)),
\end{aligned}
\end{equation}
where $$\mathcal{L}_{var}(f(x))=\log(E_u[f(x)])-E_p[\log(f(x))].$$
\subsection*{2. Proof of Eq.~\ref{eq:taylor_variational_loss_unlabeled_gradient_2}}
Given that ${0}\textless{\sum\limits_{i=1}^{n_u}f(x_i^u)}\textless{n_u}$, then let
\begin{equation}
    S=\frac{1}{n_u}\sum_{i=1}^{o}\sigma_u^{i-1},
    \label{eq:S}
\end{equation}
and
\begin{equation}
    \sigma_uS=\frac{1}{n_u}\sum_{i=1}^{o}\sigma_u^i.
    \label{eq:qS}
\end{equation}
Let Eq.~\ref{eq:S}-Eq.~\ref{eq:qS}, then
\begin{equation}\nonumber
    (1-\sigma_u)S=\frac{1}{n_u}-\frac{1}{n_u}\sigma_u^o,
\end{equation}
and then,
\begin{equation}\nonumber
    S=\frac{1-\sigma_u^o}{n_u(1-\sigma_u)},
\end{equation}
and
\begin{equation}\nonumber
    \frac{\partial \hat{\mathcal{L}}_{Tar-u}f(x)}{\partial \theta}=\frac{1-\sigma_u^o}{\sum\limits_{i=1}^{n_u}f(x_i^u)}\sum\limits_{i=1}^{n_u}\nabla_\theta f(x_i^u).
\end{equation}

\subsection*{3. Training Details for \emph{T-HOneCls}}

\paragraph{Training Details}
A detailed description of the \emph{self-calibrated optimization} is provided in Algorithm~\ref{alg:optimation}.
The hyperspectral image classification is a one-shot image input.
Stochastic gradient descent degenerates into gradient descent in the process of network optimization.
A global proportional random stratified sampler (the sampling operation in Algorithm~\ref{alg:optimation}) is also proposed to recover the stochastic gradient descent.
The detailed sampling algorithm is described in the following:
\begin{algorithm}[!h]
    \caption{Self-calibrated optimization}
    \label{alg:optimation}
    \KwIn{$H:$ hyperspectral imagery; $M_{in}:$ a set of training masks; $o:$ the order of the Taylor series; $\alpha:$ smoothing factor; $n_{pb}:$ number of pseudo batches; $T:$ training epochs; $S_{net}:$ student network; $T_{net}:$ teacher network.}
    \KwOut{The weight of the teacher network}
    \BlankLine

    Initialize the weight of the student network ($\theta_S$) and the teacher network ($\theta_T$)\\
    \For{t=1 \textbf{to} T}{
        $M_{out}=$Sampling($M_{in}, n_{pb}$)\\
        \For{e=1 \textbf{to} $n_{pb}$}{
            $p_S=S_{net}(H)$\\
            $p_T=T_{net}(H)$\\
            $\mathcal{L}_{S}=\mathcal{L}_{Tar}(p_S,M_{out}[0][e],M_{out}[1][e])$\\
            \quad $+\beta\mathcal{L}_{kl}(p_S,p_T,M_{out}[0][e],M_{out}[1][e])$\\
            update $\theta_S$\\
            update $\theta_T:\theta_T^e = \alpha\theta_T^{e-1}+(1-\alpha)\theta_S^e$
        }
    }
\end{algorithm}

\begin{algorithm}[!t]
    \caption{Global proportional random stratified sampling}
    \label{alg:sampling}
    \KwIn{$M_{in}=\left\{ m_{in}^i \right\}_{i=0}^1$: a set of training masks; $n_{pb}$:Number of pseudo batches.}
    \KwOut{$M_{out}$: a list of sets of stratified masks}
    \BlankLine

    $M_{out} \leftarrow []$ // Initialize an empty list\\
    \For{k=0 \textbf{to} 1}{
        $I_k \leftarrow \left\{j|m_{in}^{kj}=1 \right\}$\\
        $I_k \leftarrow$ Random shuffle($I_k$)\\
        $M_{out}[k] \leftarrow []$\\
        $L_k=|I_k|//n_{pb}$\\
        \While{$|I_k| \geq L_k$}{
            $r \leftarrow I_k.pop(L_k)$\\
            // Fetch $L_k$ samples from $I_k$\\
            $M_{out}[k].push(r)$\\
        }
    }
\end{algorithm}

\begin{figure}[!t]
  \centering
  \includegraphics[width=0.98\columnwidth]{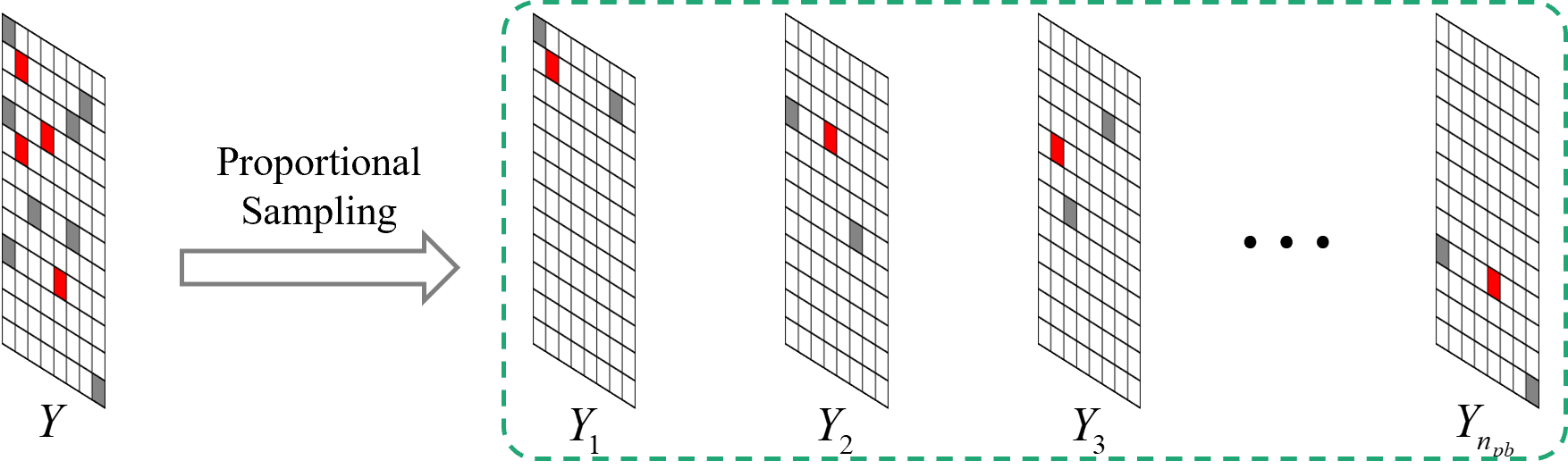}
  \caption{The description of Global proportional random stratified sampler.}
  \label{fig:sampling}
\end{figure}

\paragraph{Global Proportional Random Stratified Sampler}
Stochastic gradient descent is the mainstream optimization approach at present, so some objective functions are used based on stochastic gradient descent~\cite{nnpu,vPU,10.5555/3298483.3298518,GCE,SCE,TCE}.
Whether there will be a problem when these objective functions encounter gradient descent is not clear.
As the \emph{Taylor variational loss} can be optimized not only using stochastic gradient descent, but also using gradient descent based optimization methods, in order to ensure the adaptability of the proposed framework to different objective functions, we propose the global proportional random stratified sampler, which can recover the stochastic gradient descent by constructing pseudo-batches (such as Fig.~\ref{fig:sampling}).

The proposed sampler is summarized in Algorithm.~\ref{alg:sampling}.
The input of this sampler is a positive mask ($m_{in}^1$) and an unlabeled mask ($m_{in}^0$), and the data used for training are labeled as 1 and the other data are labeled as 0.
The key idea of the proposed sampler is to randomly train $|I_1|//n_{pb}$ positive samples and $|I_2|//n_{pb}$ unlabeled samples each time (stratified), where each batch has both positive samples and unlabeled samples (proportional).
By constructing pseudo-batches, we can meet the requirements of the current objective function for stochastic gradients.
The output of the sampler is a list of positive and unlabeled masks, with the data used for training in each batch labeled with 1 and the rest labeled with 0.

\subsection*{4. The Description of Datasets and Hyperparameters}
\begin{figure}[!t]
  \centering
  \subfloat[\small{HongHu Dataset}]{
  \label{fig:hh}
  \includegraphics[width=0.98\columnwidth]{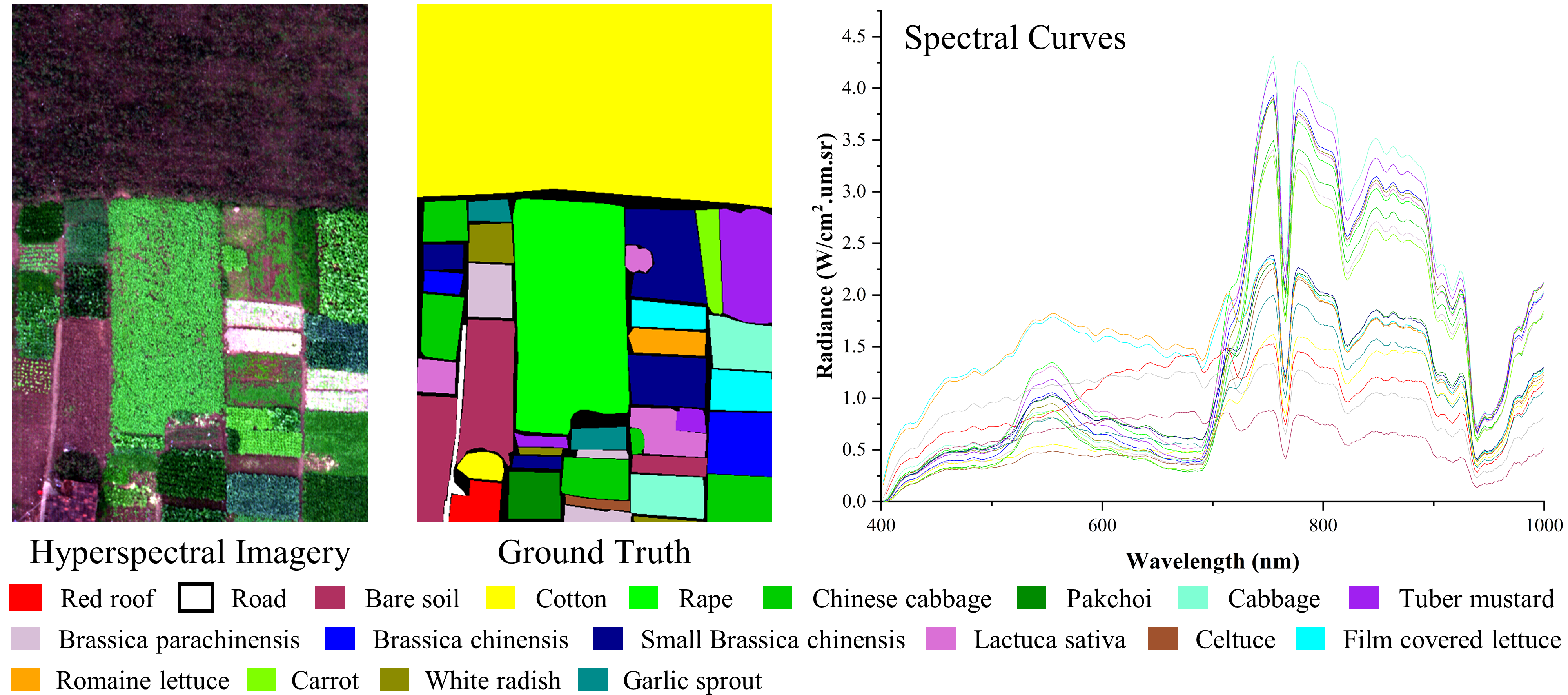}}

  \subfloat[\small{LongKou Dataset}]{
  \label{fig:lk}
  \includegraphics[width=0.98\columnwidth]{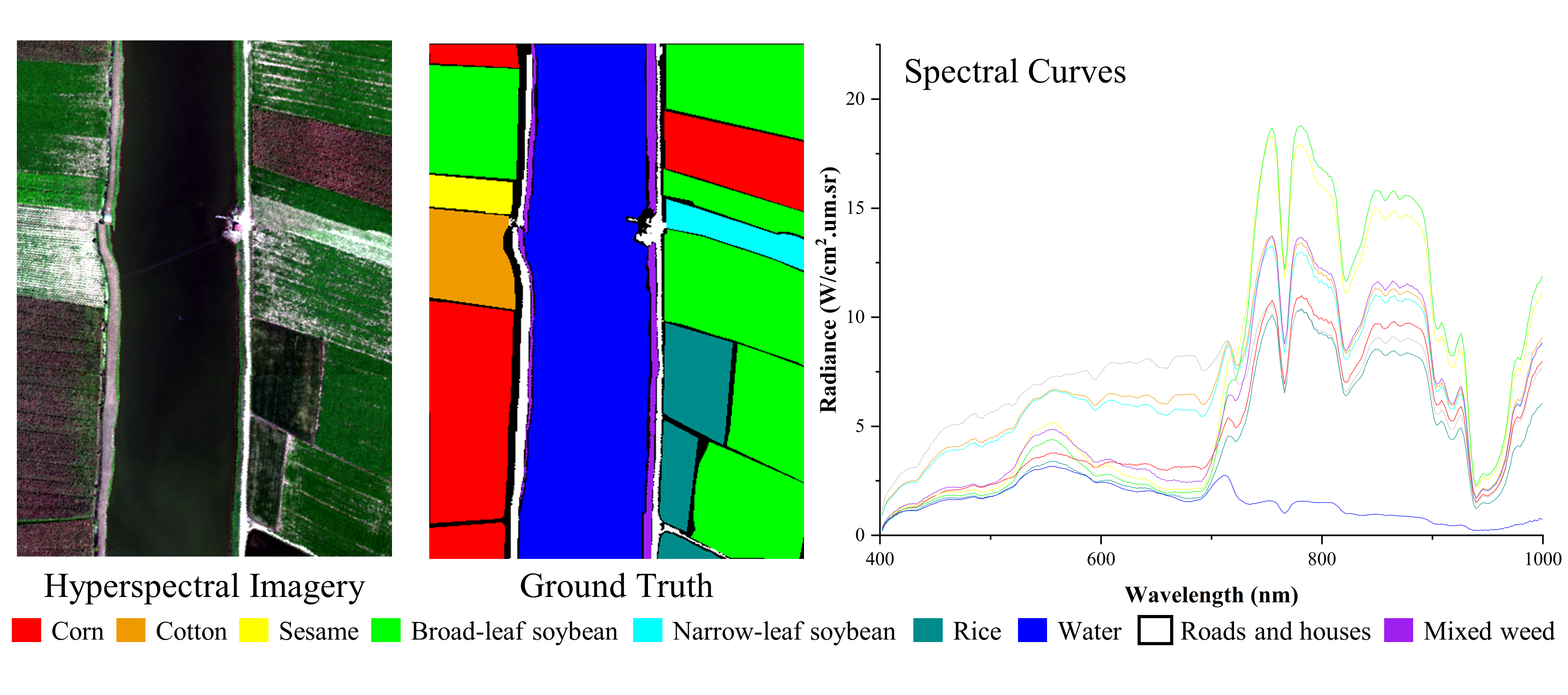}}
  
  \subfloat[\small{HanChuan Dataset}]{
  \label{fig:hc}
  \includegraphics[width=0.98\columnwidth]{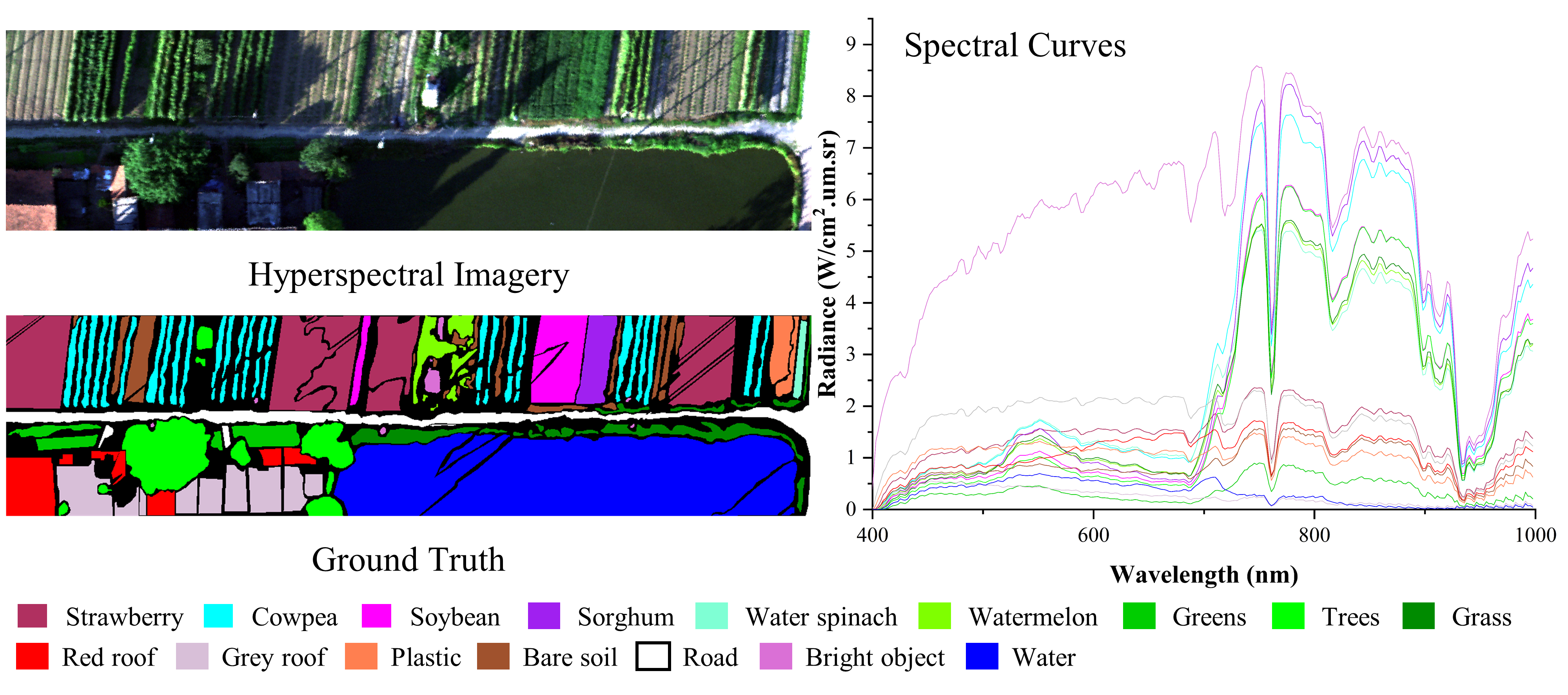}}
  \caption{UAV HSIs with ground truth and spectral curves.}
  \label{fig:hh_lk_hc_hsi_imagery}
\end{figure}

The HongHu, LongKou and HanChuan HSIs, along with the ground truth and spectral curves as examples, are shown in Fig.~\ref{fig:hh_lk_hc_hsi_imagery}.
It can be seen from the Fig.~\ref{fig:hh_lk_hc_hsi_imagery} that the spectral curves of vegetation are very similar, and it is very challenging to identify the specific vegetation types.
The hyperparameters were shown in Table~\ref{tab:hsi_details}-Table~\ref{tab:rgb_details}.

\begin{table*}[!hbt]
  \centering
  \resizebox{.98\textwidth}{!}{
    \begin{tabular}{llllll}
      \toprule
      Dataset &
        Classes selected for classification &
        \begin{tabular}[c]{@{}l@{}}Labeled samples\\ for each class\end{tabular} &
        \begin{tabular}[c]{@{}l@{}}Unlabeled samples\\ for each class\end{tabular} &
        \begin{tabular}[c]{@{}l@{}}Validation samples\\ for each class\end{tabular} &
        Hyperparameters \\ \midrule
      HongHu (270 channels) &
        \begin{tabular}[c]{@{}l@{}}Cotton, Rape,\\ Chinese cabbage,\\ Cabbage, Tuber mustard\end{tabular} &
        100 &
        4000 &
        290878 &
        \begin{tabular}[c]{@{}l@{}}Pseudo batch number: 10\\ Epochs:150\\ Optimizer: SGD (lr=0.0001, momentum=0.9, weight\_decay=0.0001)\\ with ExponentialLR (gamma=0.995)\end{tabular} \\
      LongKou (270 channels) &
        \begin{tabular}[c]{@{}l@{}}Corn, Sesame,\\ Broad-leaf soybean,\\ Rice\end{tabular} &
        100 &
        4000 &
        203642 &
        \begin{tabular}[c]{@{}l@{}}Pseudo batch number: 10\\ Epochs: 150\\ Optimizer: SGD (lr=0.0001, momentum=0.9, weight\_decay=0.0001)\\ with ExponentialLr (gamma=0.995)\end{tabular} \\
      HanChuan (274 channels) &
        \begin{tabular}[c]{@{}l@{}}Strawberry, Cowpea,\\ Soybean, Watermelon,\\ Road, Water\end{tabular} &
        100 &
        4000 &
        255930 &
        \begin{tabular}[c]{@{}l@{}}Pseudo batch number: 10\\ Epochs: 170\\ Optimizer: SGD (lr=0.0002, momentum=0.9, weight\_decay=0.0001)\\ with ExponentialLr (gamma=0.995)\end{tabular} \\ \bottomrule
      \end{tabular}}
      \caption{Details of the UAV hyperspectral datasets and hyperparameters}
      \label{tab:hsi_details}
    \end{table*}

    \begin{table*}[!hbt]
      \centering
      \resizebox{.98\textwidth}{!}{
        \begin{tabular}{llllll}
          \toprule
          Dataset                                          & Classes & \begin{tabular}[c]{@{}l@{}}Labeled samples\\ for each class\end{tabular} & \begin{tabular}[c]{@{}l@{}}Unlabeled samples\\ for each class\end{tabular} & \begin{tabular}[c]{@{}l@{}}Validation samples\\ for each class\end{tabular} & Hyperparameters                                                                                                                                                           \\ \midrule
          India Pines (200 channels)                       & 2,11    & 100                                                                      & 4000                                                                       & 10149                                                                       & \begin{tabular}[c]{@{}l@{}}Pseudo batch number:10\\ Epoch:300\\ Optimizer:SGD(lr=0.0001,momentum=0.9,weight\_decay=0.0001)\\ with ExponentialLR(gamma=0.995)\end{tabular} \\
          \multirow{2}{*}{Pavia University (103 channels)} & 2       & 100                                                                      & 4000                                                                       & 42676                                                                       & \begin{tabular}[c]{@{}l@{}}Pseudo batch number:10\\ Epoch:100\\ Optimizer:SGD(lr=0.0001,momentum=0.9,weight\_decay=0.0001)\\ with ExponentialLr(gamma=0.995)\end{tabular} \\
                                                           & 8       & 100                                                                      & 4000                                                                       & 42676                                                                       & \begin{tabular}[c]{@{}l@{}}Pseudo batch number:10\\ Epoch:300\\ Optimizer:SGD(lr=0.0001,momentum=0.9,weight\_decay=0.0001)\\ with ExponentialLr(gamma=0.995)\end{tabular} \\ \bottomrule
          \end{tabular}}
          \caption{Details of the India Pines and Pavia University datasets and hyperparameters}
          \label{tab:IP_PU_hsi_details}
        \end{table*}

        \begin{table*}[!hbt]
          \centering
          \resizebox{.98\textwidth}{!}{
            \begin{tabular}{llllll}
              \toprule
              Dataset  & Classes                                                                         & \begin{tabular}[c]{@{}l@{}}Labeled samples\\ for positive class\end{tabular} & Unlabeled samples & Validation samples & Hyperparameters                                                                              \\ \midrule
              CIFAR-10 & \begin{tabular}[c]{@{}l@{}}Positive:0,1,8,9\\ Negative:2,3,4,5,6,7\end{tabular} & 900                                                                          & 45000             & 10000              & \begin{tabular}[c]{@{}l@{}}Epoch:50\\ Order:1\\ Optimizer:Adam(lr=3e-5,betas=(0.5, 0.99))\end{tabular} \\
              STL-10   & \begin{tabular}[c]{@{}l@{}}Positive:0,2,3,8,9\\ Negative:1,4,5,6,7\end{tabular} & 900                                                                          & 99000             & 8000               & \begin{tabular}[c]{@{}l@{}}Epoch:50\\ Order:3\\ Optimizer:Adam(lr=3e-5,betas=(0.5,0.99))\end{tabular}  \\ \bottomrule
              \end{tabular}}
              \caption{Details of the CIFAR-10 and STL-10 datasets and hyperparameters}
              \label{tab:rgb_details}
            \end{table*}

\subsection*{5. The Structure of FreeOCNet}

\begin{figure}[!h]
    \centering
    \includegraphics[width=0.98\columnwidth]{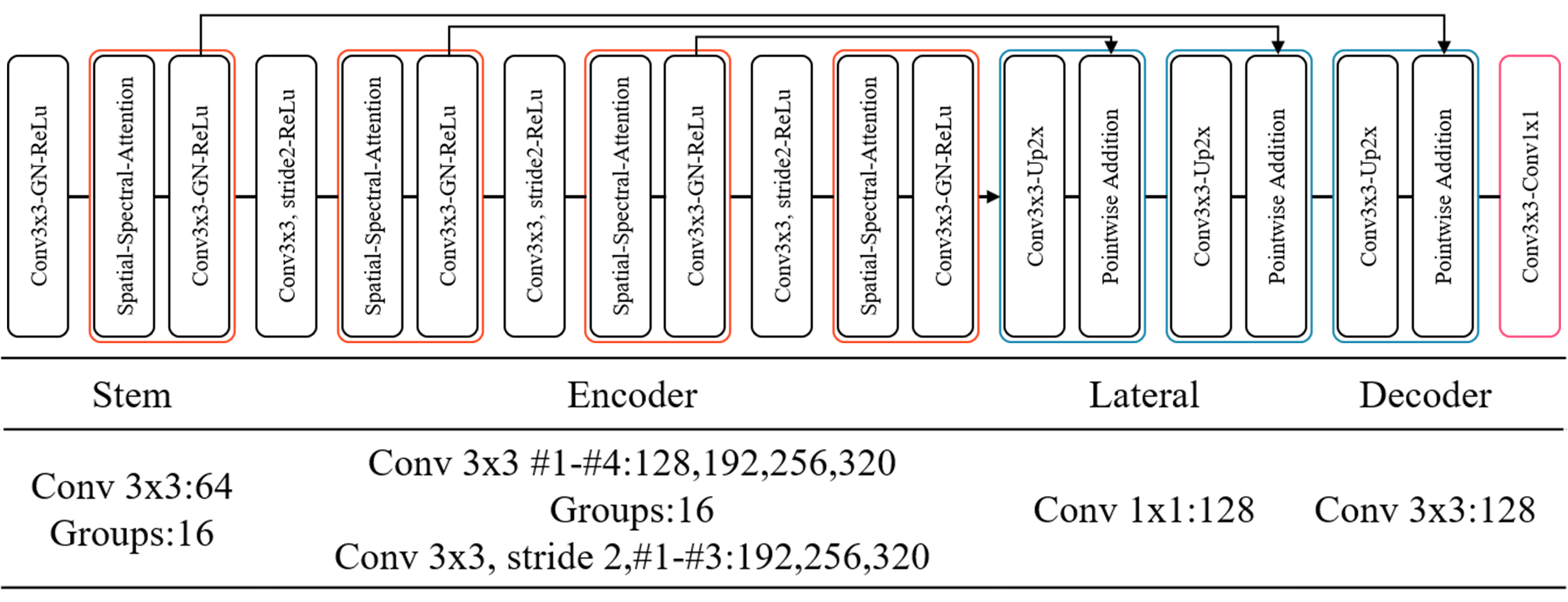}
    \caption{The description of FreeOCNet.}
    \label{fig:freeocnet}
\end{figure}

The FreeOCNet includes encoder, decoder and lateral connection (Fig.~\ref{fig:freeocnet}).
The basic module in encoder is a spectral-spatial-attention (SSA)-convolution layer (Conv $3\time3$)-Group normalization-rectified linear unit (ReLU), and the mdoule of a Conv $3\times3$ with stride 2 to reduce the spatial size.
A lightweight decoder is used, which consists of a Conv $3\times3$ layer and $2\times$ upsampling layer and a fixed number of channels.
A Conv $1\times1$ layer is used in lateral connection to reduce the number of channels in the encoder.

\subsection*{6. More Experimental Results}

\begin{figure*}[!t]
\centering
\subfloat[\small{Distribution maps for the HongHu dataset.}]{
\label{fig:honghu_distribution_maps}
\includegraphics[width=0.9\textwidth]{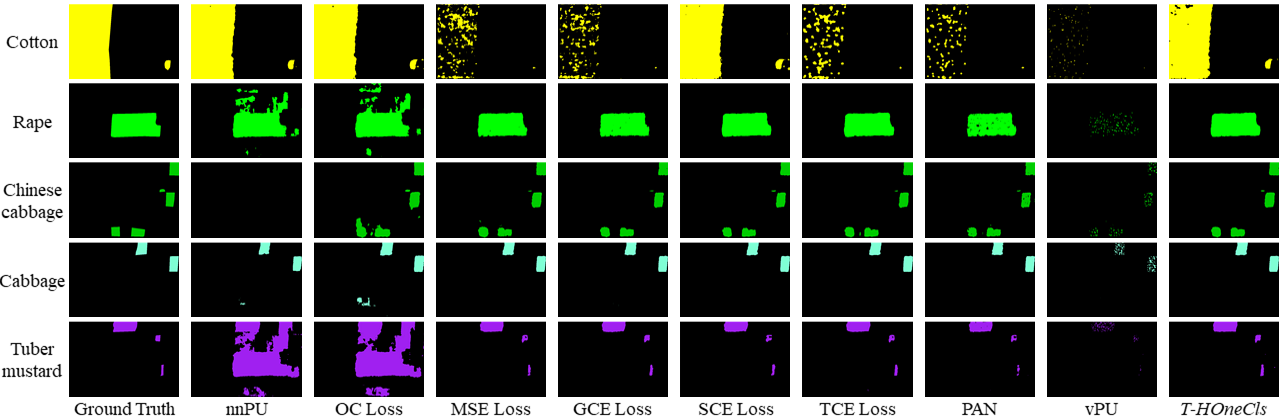}}

\subfloat[\small{Distribution maps for the LongKou dataset.}]{
\label{fig:longkou_distribution_maps}
\includegraphics[width=0.9\textwidth]{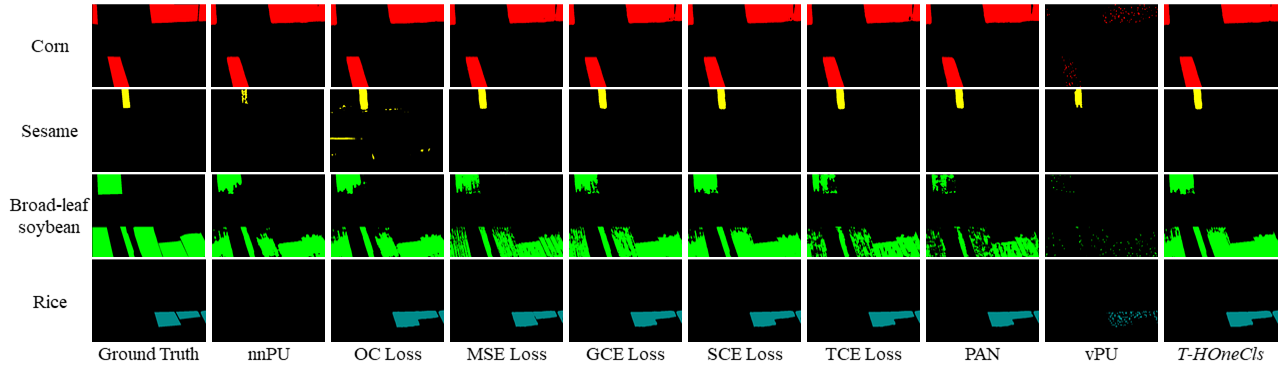}}

\subfloat[\small{Distribution maps for the HanChuan dataset.}]{
\label{fig:hanchuan_distribution_maps}
\includegraphics[width=0.9\textwidth]{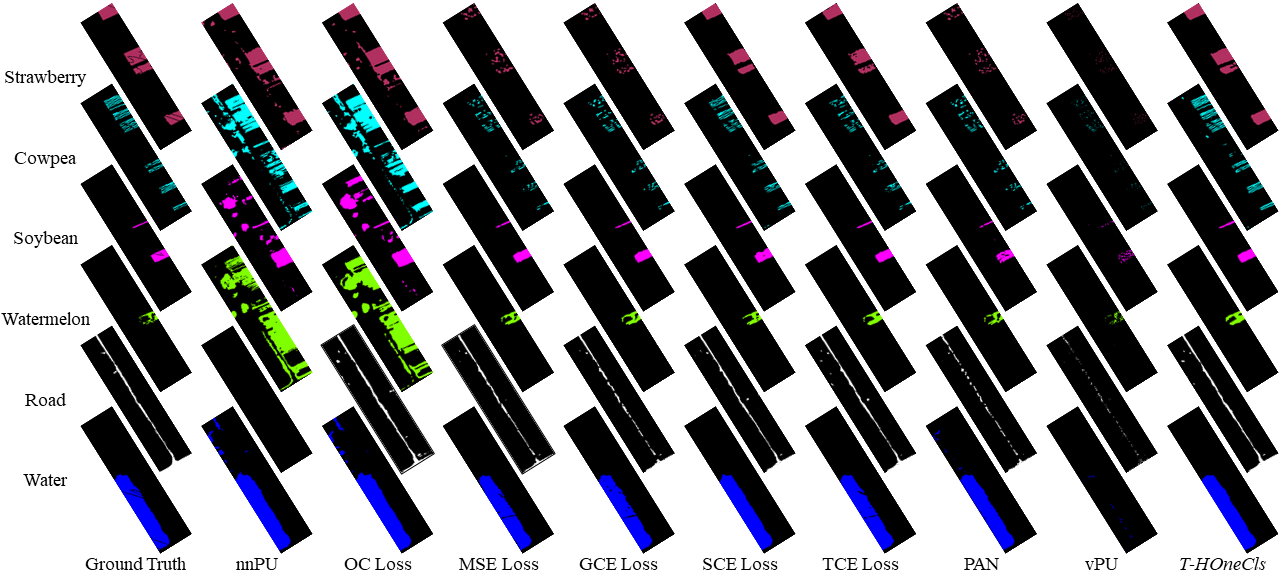}}
\caption{Distribution maps for the UAV hyperspectral datasets. The maps with the best F1-score are displayed for five experiments.}
\label{fig:distribution_maps}
\end{figure*}

The distribution maps for the HongHu, LongKou and HanChuan datasets are shown in Fig.~\ref{fig:honghu_distribution_maps}, Fig.~\ref{fig:longkou_distribution_maps} and Fig.~\ref{fig:hanchuan_distribution_maps}, respectively.
The Precision and Recall for the HongHu, HanChuan and LongKou datasets are shown in Table~\ref{tab:HongHu_pr}, Table~\ref{tab:HanChuan_pr} and Table~\ref{tab:LongKou_pr}, respectively.
As shown in this subsection, other methods cannot obtain high precision and recall at the same time, that is, these methods cannot find a balance between the overfitting and underfitting of the training data.
This balance was found by \textit{T-HOneCls}, and a good F1-score was obtained by \textit{T-HOneCls} in all tasks.

\begin{table*}[!hbt]
  \centering
  \resizebox{.98\textwidth}{!}{
    \begin{tabular}{cccccccccc}
      \toprule
      \multicolumn{1}{c|}{\multirow{2}{*}{Class}} & \multicolumn{2}{c|}{Class prior-based classifiers} & \multicolumn{4}{c|}{Label noise representation learning}                 & \multicolumn{3}{c}{Class prior-free classifiers} \\
      \multicolumn{1}{c|}{}                       & nnPU            & \multicolumn{1}{c|}{OC Loss}     & MSE Loss    & GCE Loss     & SCE Loss    & \multicolumn{1}{c|}{TCE Loss} & PAN            & vPU            & T-HOneCls      \\ \midrule
      Cotton                                      & 99.34/99.54     & 99.31/\textbf{99.57}                      & 99.98/9.52  & \textbf{100.00}/10.19 & 99.94/93.08 & \textbf{100.00}/11.29                  & \textbf{100.00}/9.09    & 99.94/0.94     & 99.96/96.40    \\
      Rape                                        & 69.79/99.58     & 69.38/\textbf{99.71}                      & 99.86/93.03 & 99.86/93.73  & 99.88/94.95 & 99.78/95.59                   & \textbf{99.93}/64.81    & 99.74/4.34     & 99.77/95.92    \\
      Chinese cabbage                             & 0.00/0.00       & 81.19/\textbf{96.29}                      & 95.96/91.40 & 97.81/90.60  & 97.58/90.27 & 97.31/91.27                   & \textbf{98.19}/87.11    & 97.01/14.28    & 95.97/92.60    \\
      Cabbage                                     & 54.18/54.48     & 81.55/\textbf{99.91}                      & 99.87/98.54 & 99.89/98.32  & 99.89/98.37 & 99.85/98.75                   & \textbf{99.92}/96.50    & 99.88/21.12    & 99.79/98.95    \\
      Tuber mustard                               & 13.63/99.73     & 13.36/\textbf{99.88}                     & 99.00/91.75 & 98.68/93.57  & 98.72/92.49 & 98.41/94.87                   & \textbf{99.33}/86.00    & 99.30/13.19    & 98.56/96.24    \\ \bottomrule
      \end{tabular}}
    \caption{The Precision/Recall for the HongHu dataset}
    \label{tab:HongHu_pr}
\end{table*}

\begin{table*}[!hbt]
  \centering
  \resizebox{.98\textwidth}{!}{
    \begin{tabular}{cccccccccc}
      \toprule
      \multicolumn{1}{c|}{\multirow{2}{*}{Class}} & \multicolumn{2}{c|}{Class prior-based classifiers} & \multicolumn{4}{c|}{Label noise representation learning}                & \multicolumn{3}{c}{Class prior-free classifiers} \\
      \multicolumn{1}{c|}{}                       & nnPU            & \multicolumn{1}{c|}{OC Loss}     & MSE Loss    & GCE Loss    & SCE Loss    & \multicolumn{1}{c|}{TCE Loss} & PAN            & vPU            & T-HOneCls      \\ \midrule
      Strawberry                                  & 80.94/99.30     & 81.22/\textbf{99.75}                      & \textbf{99.85}/20.38 & 99.76/20.93 & 99.78/86.12 & 99.81/66.12                   & 99.74/18.32    & 98.50/4.94     & 99.16/90.44    \\
      Cowpea                                      & 42.79/\textbf{98.91}     & 42.12/98.61                      & 99.83/30.40 & \textbf{99.94}/30.13 & 98.90/55.82 & 99.83/39.77                   & 99.89/31.37    & 99.04/6.86     & 96.69/84.77    \\
      Soybean                                     & 27.96/99.85     & 26.86/\textbf{99.98}                      & \textbf{99.69}/95.27 & 99.51/95.13 & 99.43/95.07 & 99.56/97.55                   & 98.93/77.48    & 96.86/24.22    & 99.62/98.64    \\
      Watermelon                                  & 6.25/\textbf{99.97}      & 6.52/99.76                       & 93.21/94.89 & 94.14/93.48 & 93.45/93.50 & 91.00/94.42                   & 96.72/87.70    & \textbf{98.31}/38.00    & 89.23/97.11    \\
      Road                                        & 0.00/0.00       & 86.81/\textbf{92.48}                      & 98.71/62.71 & 98.50/60.07 & 96.65/77.07 & 98.65/76.78                   & \textbf{99.46}/44.60    & 98.48/14.34    & 97.73/86.43    \\
      Water                                       & 90.99/99.95     & 90.34/\textbf{99.96}                      & 98.67/79.54 & 99.67/85.96 & 99.58/94.50 & 98.73/90.42                   & 99.54/62.69    & 99.76/0.72     & \textbf{100.00}/96.79   \\ \bottomrule
      \end{tabular}}
    \caption{The Precision/Recall for the HanChuan dataset}
    \label{tab:HanChuan_pr}
\end{table*}

\begin{table*}[!hbt]
  \centering
  \resizebox{.98\textwidth}{!}{
    \begin{tabular}{cccccccccc}
      \toprule
      \multicolumn{1}{c|}{\multirow{2}{*}{Class}} & \multicolumn{2}{c|}{Class prior-based classifiers} & \multicolumn{4}{c|}{Label noise representation learning}                & \multicolumn{3}{c}{Class prior-free classifiers} \\
      \multicolumn{1}{c|}{}                       & nnPU            & \multicolumn{1}{c|}{OC Loss}     & MSE Loss    & GCE Loss    & SCE Loss    & \multicolumn{1}{c|}{TCE Loss} & PAN            & vPU             & T-HOneCls     \\ \midrule
      Corn                                        & 99.89/97.29     & 99.48/\textbf{99.87}                      & 99.96/98.92 & 99.94/98.38 & 99.98/97.08 & 99.96/97.71                   & 99.95/94.59    & \textbf{100.00}/4.46     & 99.92/99.49   \\
      Sesame                                      & 20.00/7.55      & 61.29/\textbf{100.00}                     & 99.93/99.61 & \textbf{99.97}/99.58 & \textbf{99.97}/99.59 & 99.91/99.67                   & 99.94/99.53    & \textbf{99.97}/53.04     & 99.94/99.70   \\
      Broad-leaf soybean                          & 98.69/74.19     & 96.10/81.23                      & \textbf{99.98}/69.56 & 99.92/77.53 & 99.93/77.20 & \textbf{99.98}/60.04                   & \textbf{99.98}/41.35    & 99.93/2.28      & 99.88/\textbf{86.39}   \\
      Rice                                  & 0.00/0.00       & 99.41/\textbf{100.00}                     & 99.96/97.95 & 99.96/98.43 & 99.98/98.36 & 99.97/97.64                   & 99.99/97.30    & \textbf{100.00}/21.17    & 99.96/99.04   \\ \bottomrule
      \end{tabular}}
    \caption{The Precision/Recall for the LongKou dataset}
    \label{tab:LongKou_pr}
\end{table*}

\begin{figure*}[!hbt]
  \centering
  \subfloat[\small{Positive loss of the variational classifier}]{
  \label{fig:positive_vpu_loss_6}
  \includegraphics[width=0.3\textwidth]{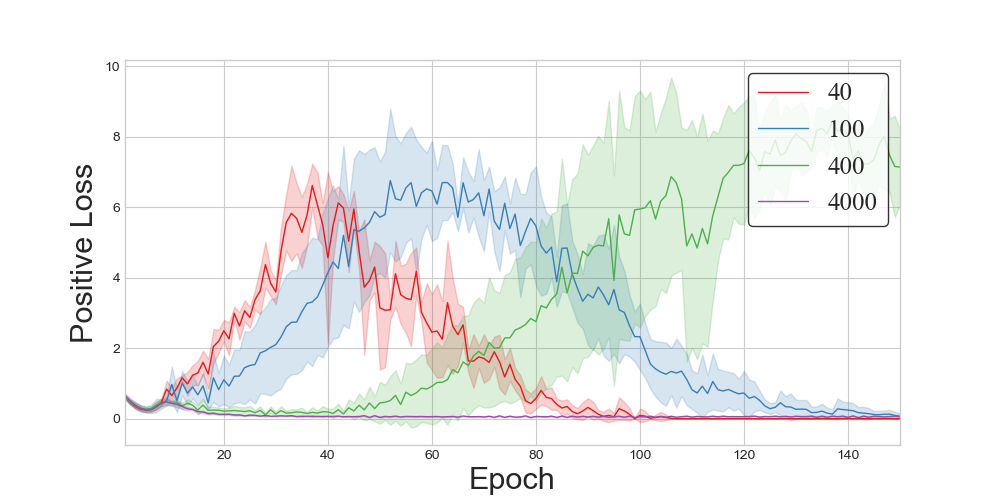}}
  \subfloat[\small{Total loss of the variational classifier}]{
  \label{fig:total_vpu_loss_6}
  \includegraphics[width=0.3\textwidth]{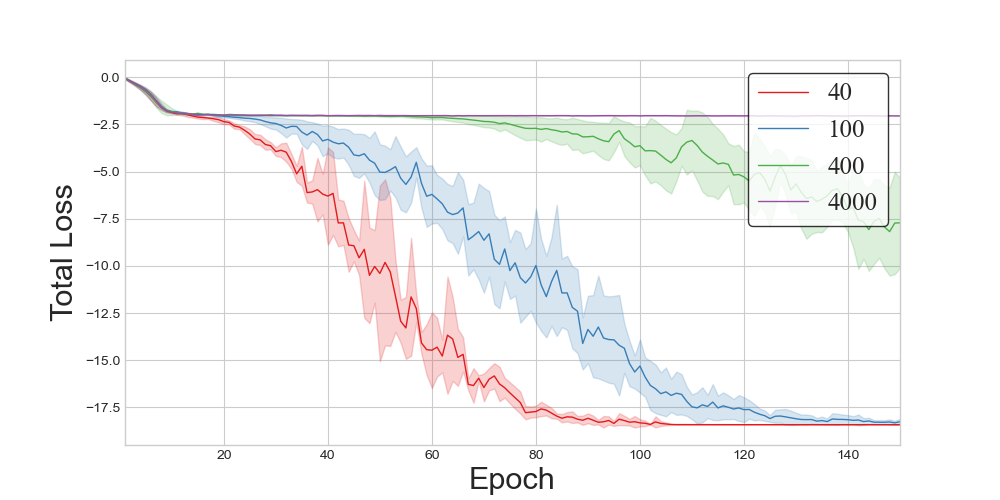}}
  \subfloat[\small{F1-score of the variational classifier}]{
  \label{fig:f1_vpu_6}
  \includegraphics[width=0.3\textwidth]{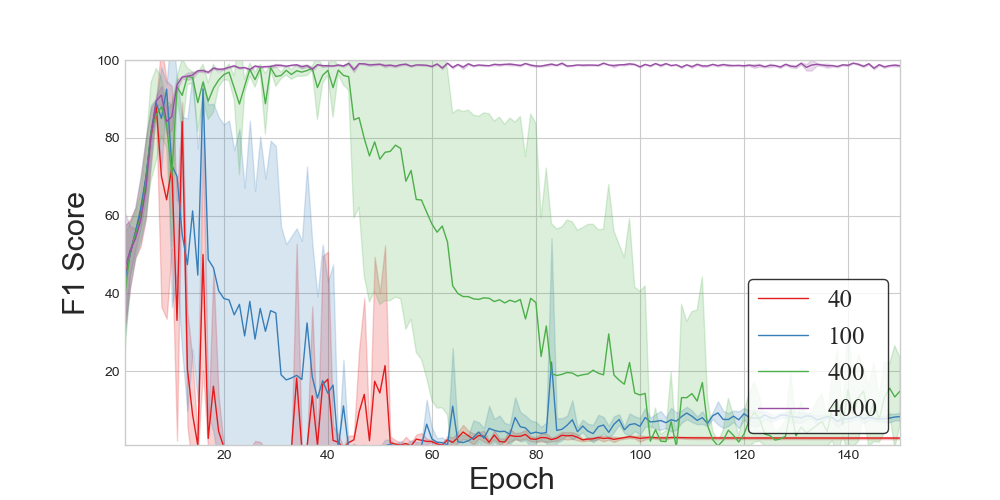}}

  \subfloat[\small{Positive loss of \emph{T-HOneCls}}]{
  \label{fig:positive_t_loss_6}
  \includegraphics[width=0.3\textwidth]{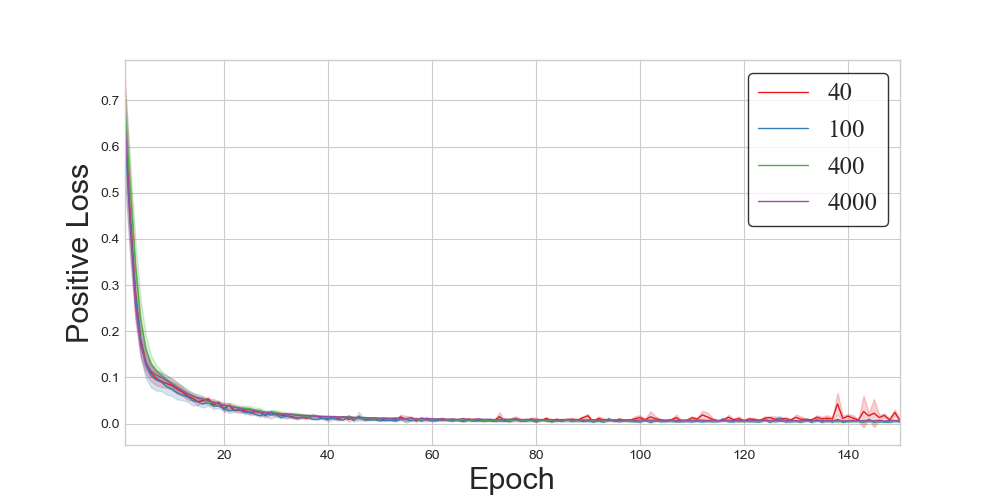}}
  \subfloat[\small{Total loss of \emph{T-HOneCls}}]{
  \label{fig:total_t_loss_6}
  \includegraphics[width=0.3\textwidth]{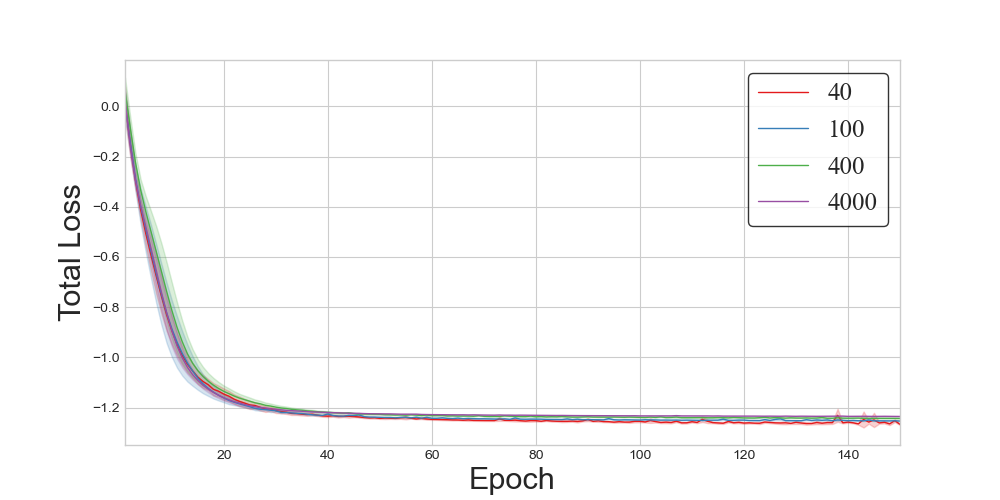}}
  \subfloat[\small{F1-score of \emph{T-HOneCls}}]{
  \label{fig:f1_t_6}
  \includegraphics[width=0.3\textwidth]{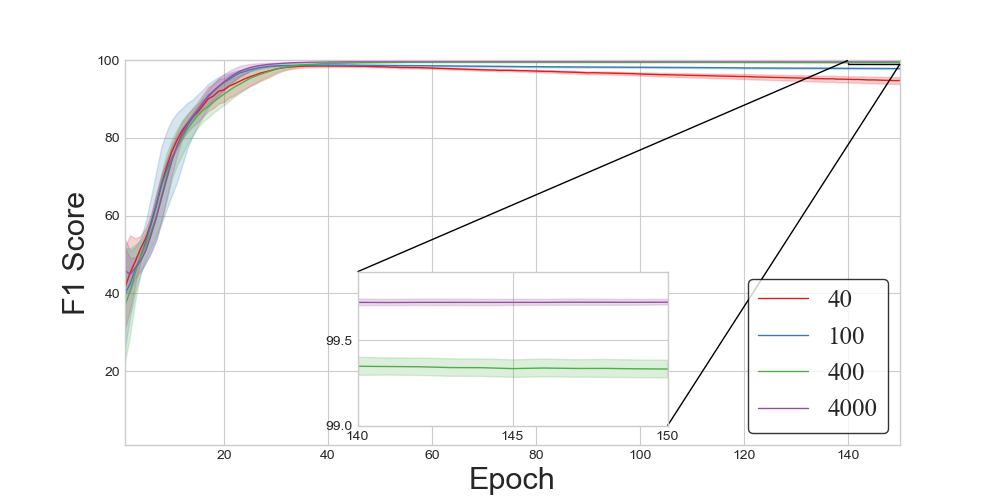}}
  \caption{The curves of rape in the HongHu dataset, showing the positive loss, total loss, and F1-score of the variational classifier and \emph{T-HOneCls} with different positive training samples in the training stage.}
  \label{fig:loss_and_f1_6}
\end{figure*}

\begin{figure*}[!hbt]
  \centering
  \subfloat[\small{Positive loss of the variational classifier}]{
  \label{fig:positive_vpu_loss_9}
  \includegraphics[width=0.3\textwidth]{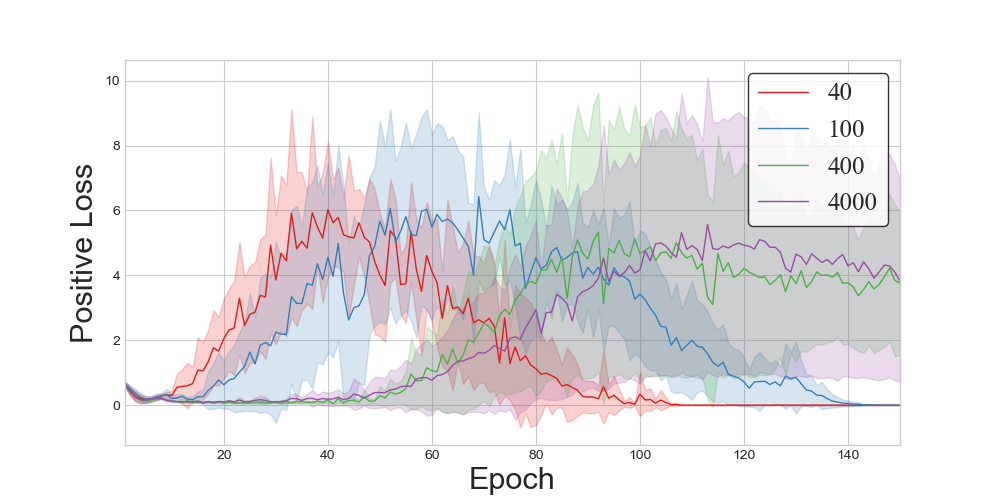}}
  \subfloat[\small{Total loss of the variational classifier}]{
  \label{fig:total_vpu_loss_9}
  \includegraphics[width=0.3\textwidth]{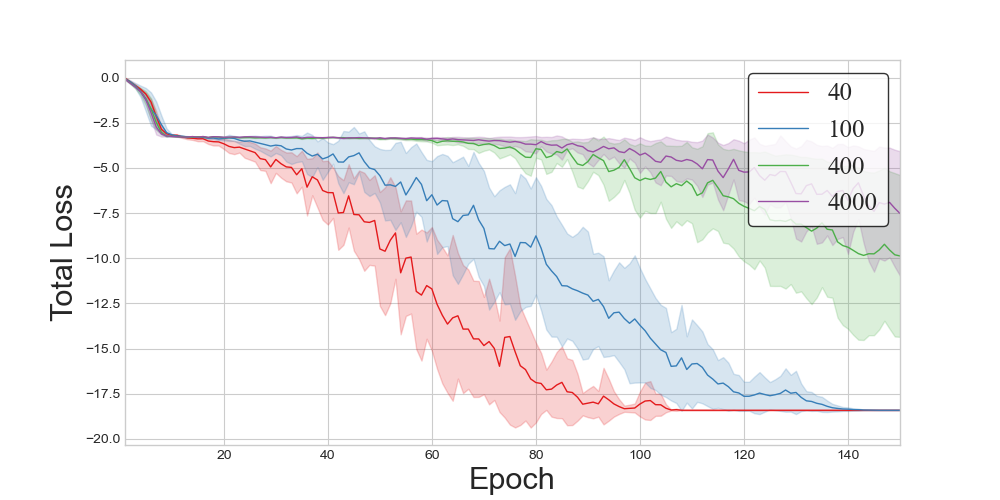}}
  \subfloat[\small{F1-score of the variational classifier}]{
  \label{fig:f1_vpu_9}
  \includegraphics[width=0.3\textwidth]{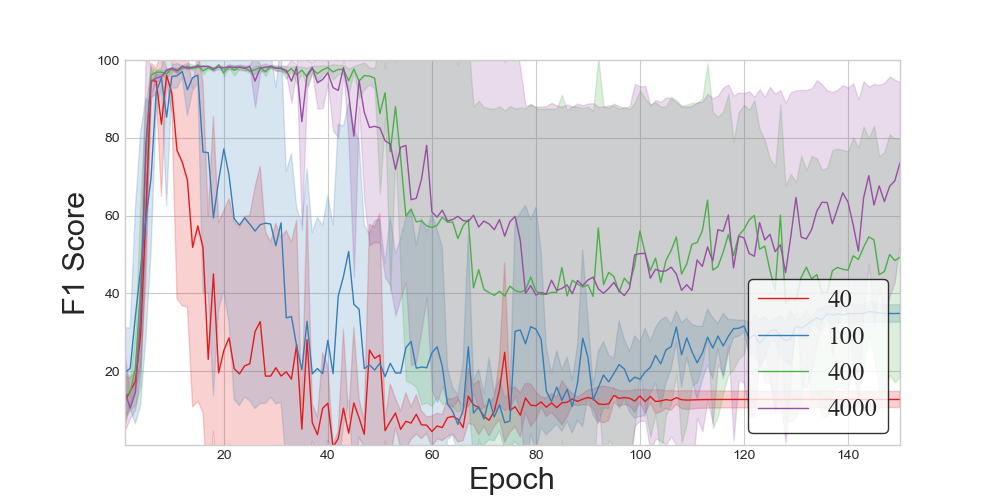}}

  \subfloat[\small{Positive loss of \emph{T-HOneCls}}]{
  \label{fig:positive_t_loss_9}
  \includegraphics[width=0.3\textwidth]{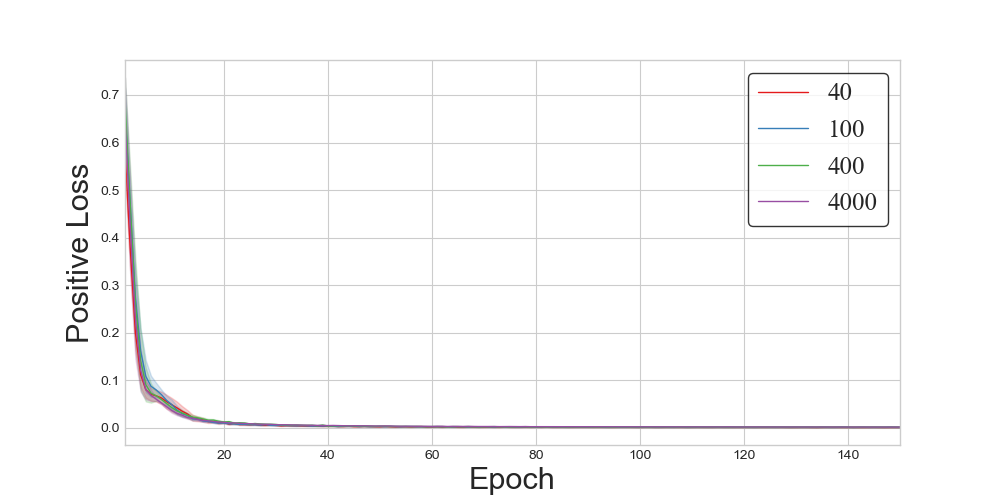}}
  \subfloat[\small{Total loss of \emph{T-HOneCls}}]{
  \label{fig:total_t_loss_9}
  \includegraphics[width=0.3\textwidth]{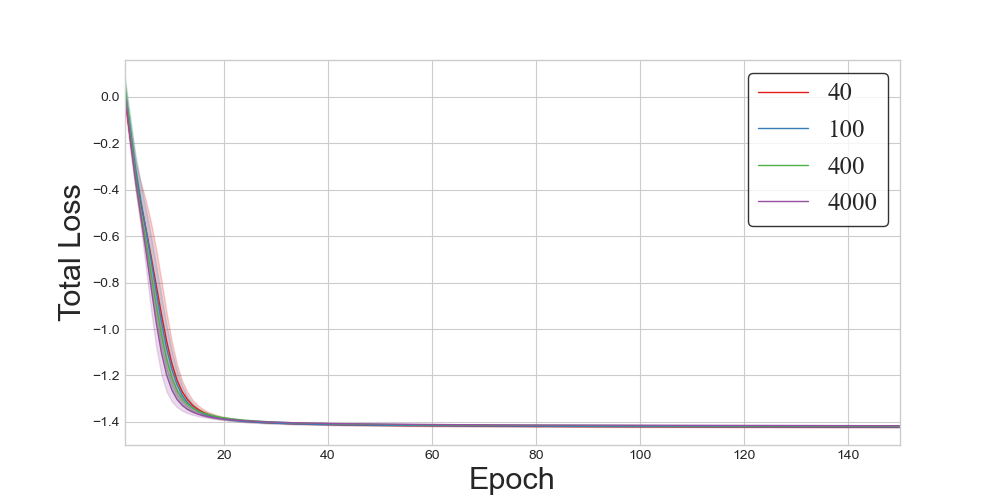}}
  \subfloat[\small{F1 score of \emph{T-HOneCls}}]{
  \label{fig:f1_t_9}
  \includegraphics[width=0.3\textwidth]{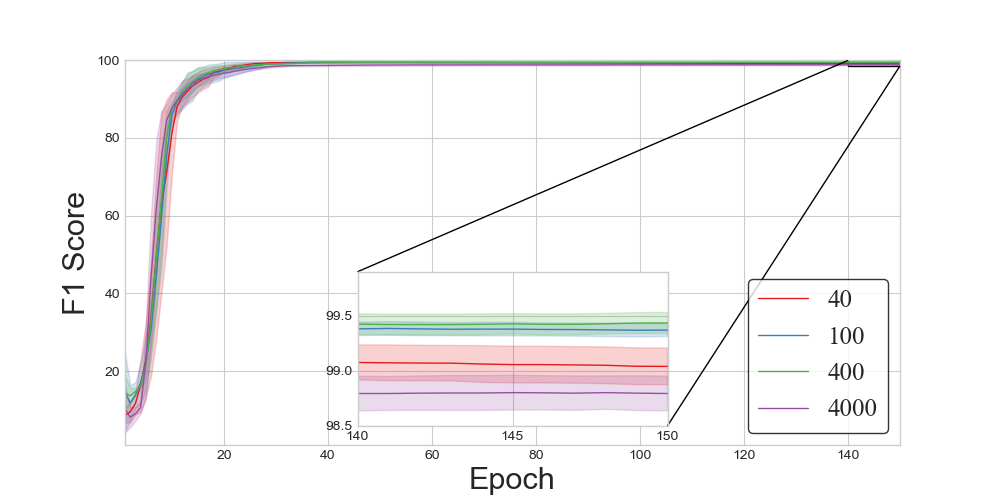}}
  \caption{The curves of cabbage in the HongHu dataset, showing the positive loss, total loss and F1-score of the variational classifier and \emph{T-HOneCls} with different positive training  samples in the training stage.}
  \label{fig:loss_and_f1_9}
\end{figure*}

\subsection*{7. More Experimental Results for the Training Process and Training Samples}

The curves of the positive class and the total loss of the different positive training samples of rape and cabbage are shown in Fig.~\ref{fig:loss_and_f1_6} and Fig.~\ref{fig:loss_and_f1_9}, respectively.
The curves of the F1-score are also shown.
The variational loss using fewer training samples leads to the gradient domination optimization process of unlabeled samples at the beginning of the training, which makes the loss of positive classes rise at the beginning of the training.
Although the loss of the positive samples will decreases as the training progresses, for example 40, 100 or 400, the F1-score is unstable, and determining the optimal training epochs is very challenging without using additional data.
In the classification of rape in the HongHu dataset, 4000 positive training samples can obtain a stable F1-score, but the F1-score of cabbage is still unstable.
However, this shortcoming is overcome by the proposed \emph{T-HOneCls}, and a stable F1-score can be obtained, as shown in Fig.~\ref{fig:f1_t_6} and Fig.~\ref{fig:f1_t_9}.

\begin{figure*}[!hbt]
  \centering
  \subfloat[\small{Cotton in HongHu}]{
  \label{fig:cotton_in_hh_order}
  \includegraphics[width=0.3\textwidth]{Figures/order_experiments/HongHu_4.png}}
  \subfloat[\small{Rape in HongHu}]{
  \label{fig:rape_in_hh_order}
  \includegraphics[width=0.3\textwidth]{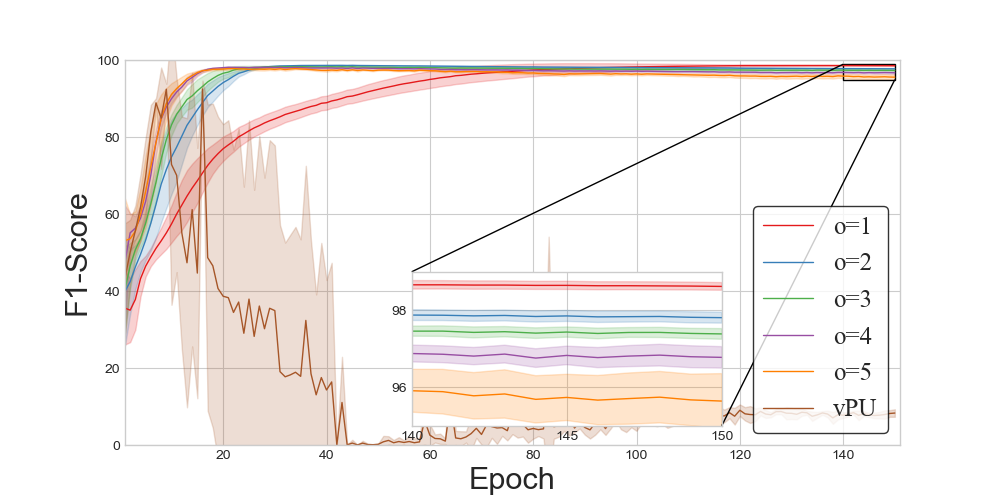}}
  \subfloat[\small{Sesame in LongKou}]{
  \label{fig:sesame_in_lk_order}
  \includegraphics[width=0.3\textwidth]{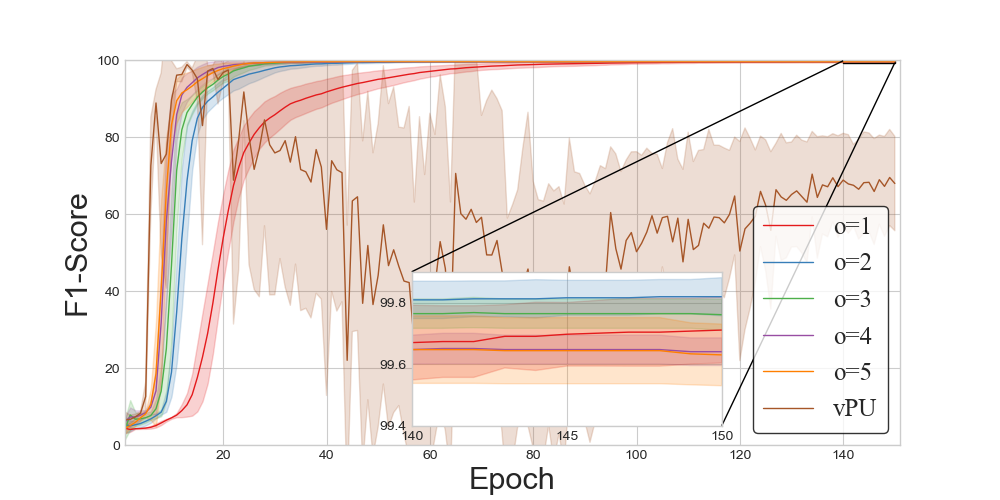}}
  
  \subfloat[\small{Broad-leaf soybean in LongKou}]{
  \label{fig:broad-leaf_sotbean_in_lk_order}
  \includegraphics[width=0.3\textwidth]{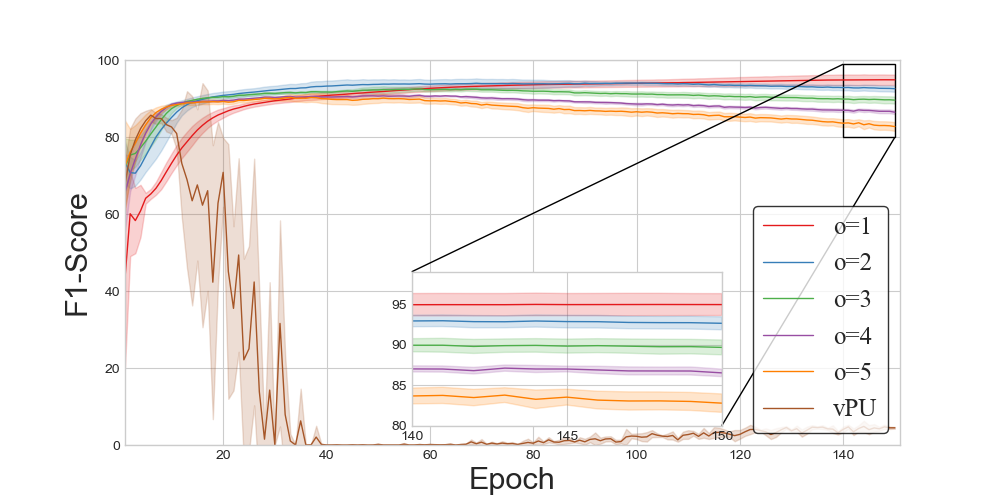}}
  \subfloat[\small{Cowpea in HanChuan}]{
  \label{fig:cowpea_in_hc_order}
  \includegraphics[width=0.3\textwidth]{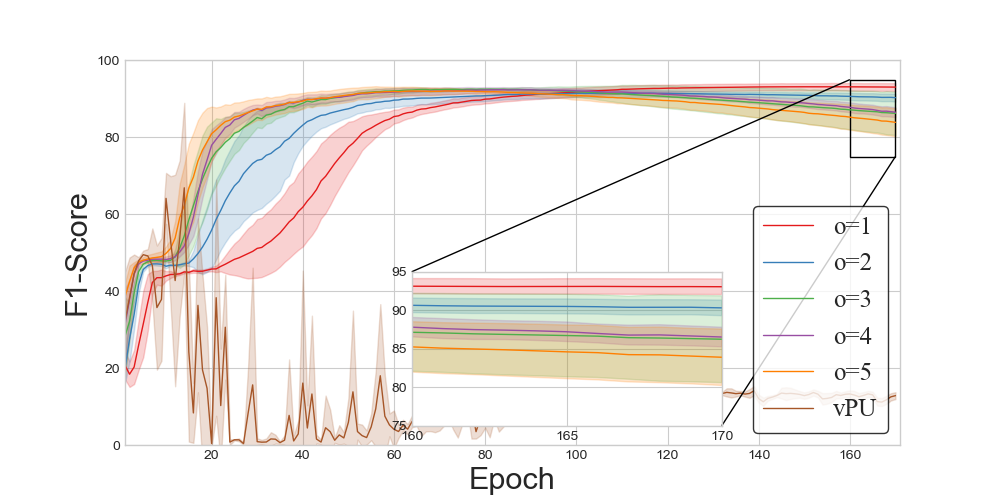}}
  \subfloat[\small{Watermelon in HanChuan}]{
  \label{fig:watermelon_in_hc_order}
  \includegraphics[width=0.3\textwidth]{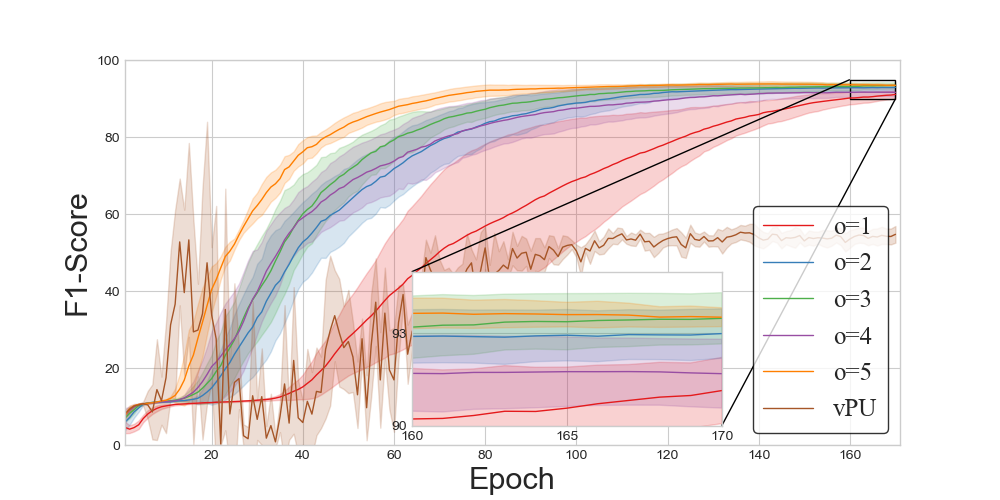}}
  \caption{The F1-score curves for the different order of the Taylor series in \emph{T-HOneCls}.}
  \label{fig:order_experiments_2}
\end{figure*}

\begin{figure*}[!hbt]
  \centering
  \subfloat[\small{Cotton in HongHu (o=2)}]{
  \label{fig:cotton_in_hh_kl_2}
  \includegraphics[width=0.3\textwidth]{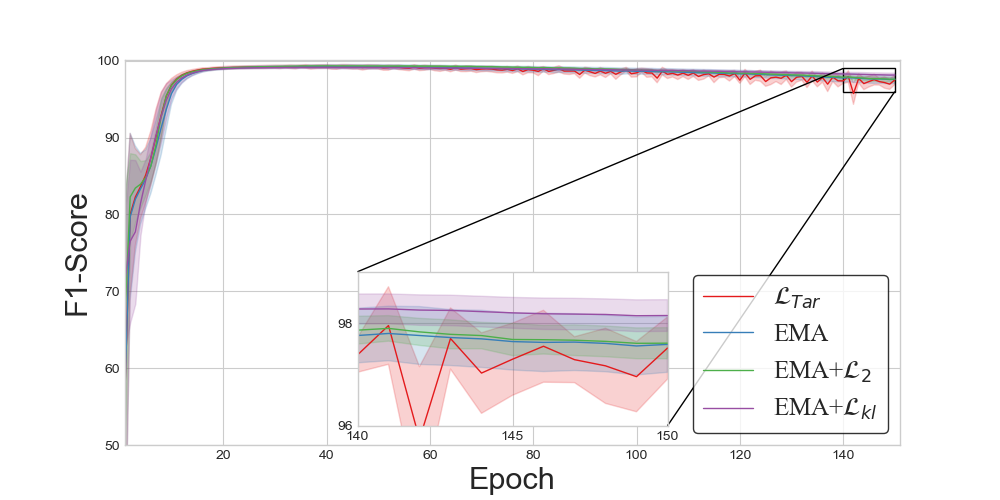}}
  \subfloat[\small{Broad-leaf soybean in LongKou (o=2)}]{
  \label{fig:broad-leaf_sotbean_in_lk_kl_2}
  \includegraphics[width=0.3\textwidth]{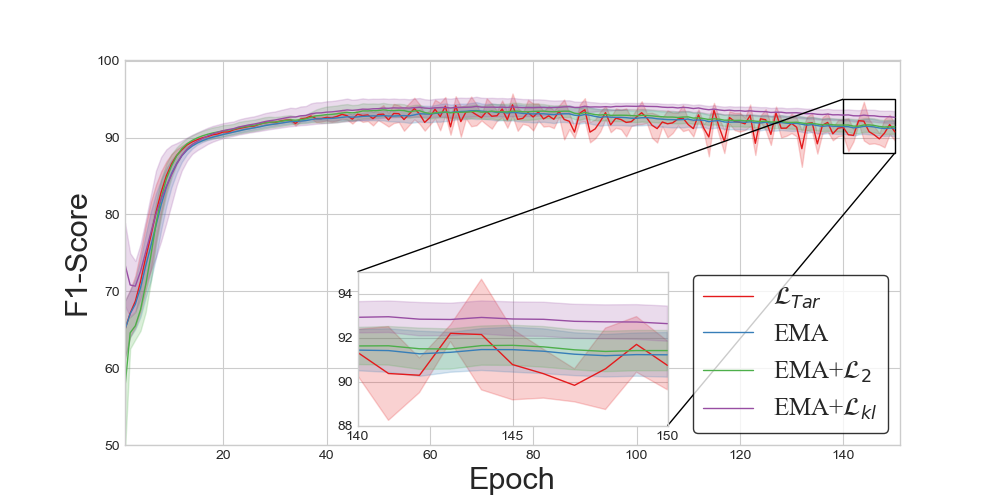}}
  \subfloat[\small{Cowpea in HanChuan (o=2)}]{
  \label{fig:watermelon_in_hc_kl_2}
  \includegraphics[width=0.3\textwidth]{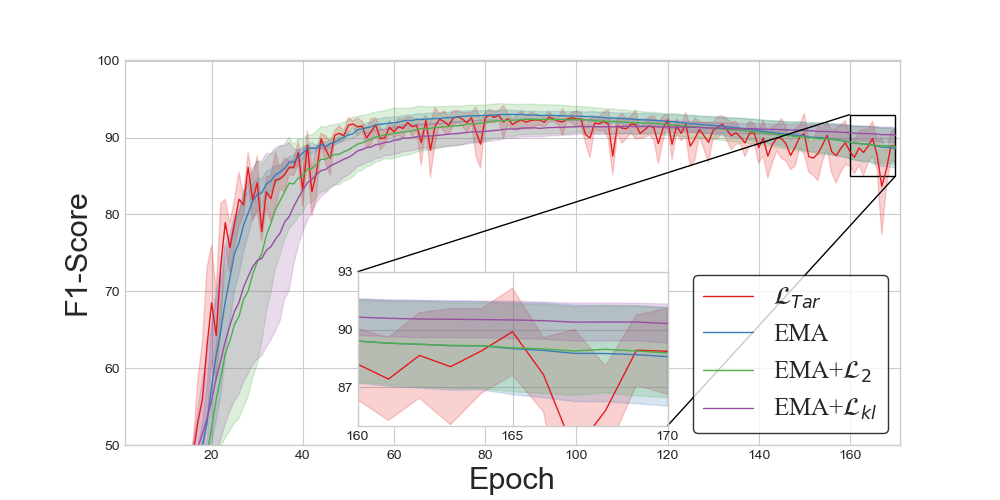}}

  \subfloat[\small{Cotton in HongHu (o=5)}]{
  \label{fig:cotton_in_hh_kl_5}
  \includegraphics[width=0.3\textwidth]{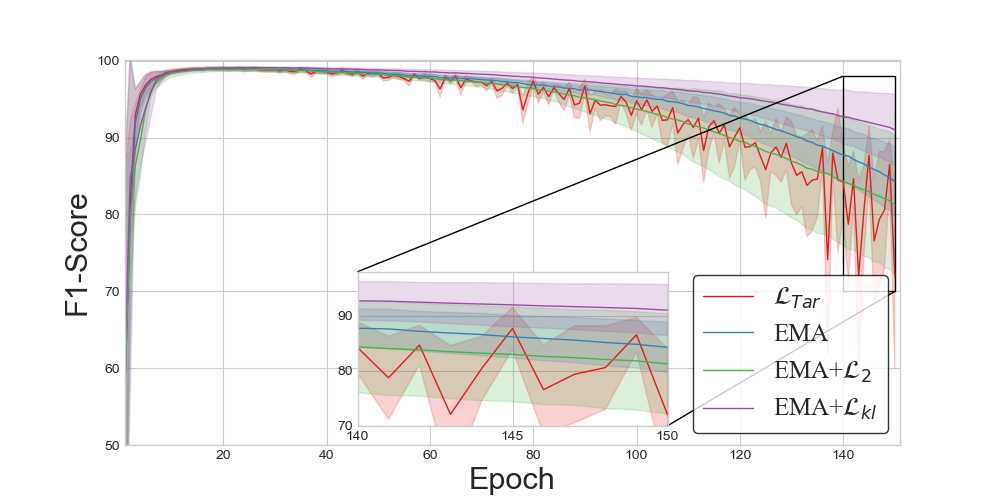}}
  \subfloat[\small{Broad-leaf soybean in LongKou (o=5)}]{
  \label{fig:broad-leaf_sotbean_in_lk_kl_5}
  \includegraphics[width=0.3\textwidth]{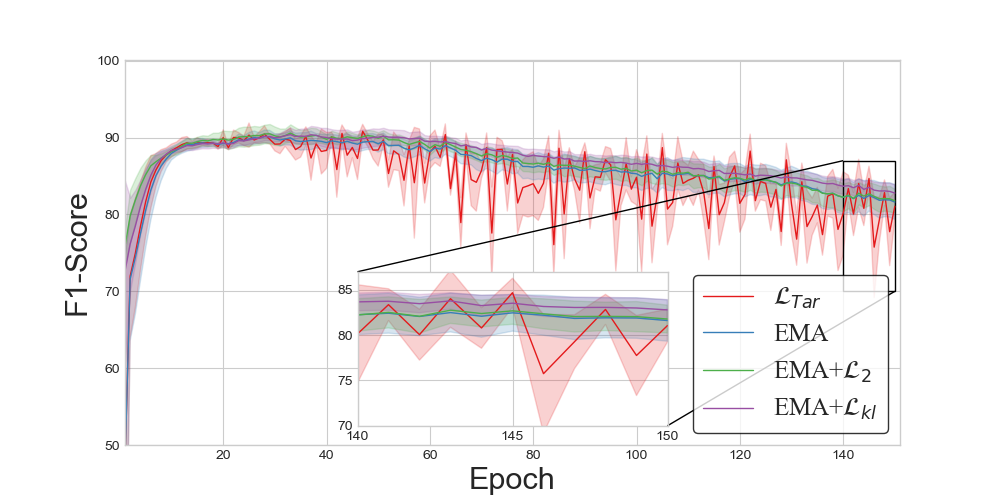}}
  \subfloat[\small{Cowpea in HanChuan (o=5)}]{
  \label{fig:watermelon_in_hc_kl_5}
  \includegraphics[width=0.3\textwidth]{Figures/kl-teacher_ecperiments/HanChuan_2-5.png}}
  \caption{The F1-score curves of the different components of KL-Teacher in \emph{T-HOneCls}.}
  \label{fig:kl_teacher_experiments_2}
\end{figure*}

\subsection*{8. More Experimental Results for the Order of the Taylor Series}

The results for cotton and five other ground objects are displayed in Fig.~\ref{fig:order_experiments_2}.
The most important contribution of this paper is to point out that the reason for the poor performance of variational loss is that the gradient of the unlabeled data is given too much weight.
To solve this problem, Taylor expansion is introduced in the variational loss, so as to reduce the weight of the unlabeled data in the gradient.
An empirical conclusion can be obtained from Fig.~\ref{fig:order_experiments_2}: the higher the order of the Taylor expansion, the faster the neural network converges.
However, the rapid convergence of the neural network can lead to overfitting.
In other words, the classification results first rise and then decline with the progress of the training, as shown in the curve of $o=5$ in Fig.~\ref{fig:cotton_in_hh_order}.
A small expansion order slows down the convergence of the neural network, as shown in the curve of $o=1$ in Fig.~\ref{fig:watermelon_in_hc_order}.
Empirically, a higher Taylor expansion order can be equipped with fewer training epochs, and a smaller Taylor expansion order can be equipped with more training epochs.
In order to show that \emph{T-HOneCls} can significantly reduce the overfitting of the neural network for noisy labels, we set a relatively large number of training epochs, so that $o\in\{1,2,3,4\}$ can achieve s good F1-score. Finally, we set $o=2$.

\subsection*{9. More Experimental Results about KL-Teacher}

The results of other classes are shown in Fig.~\ref{fig:kl_teacher_experiments_2}.
The first thing to be analyzed is the role of EMA in the self-calibration optimization.
It can be seen from Fig.~\ref{fig:kl_teacher_experiments_2} shows that the F1-score fluctuates greatly when only stochastic gradient descent is used to optimize the \emph{Taylor variational loss}, and in this case, selecting appropriate training epochs can seriously affect the F1-score of the model.
EMA has the function of an ``F1-score filter'', which makes the F1-score of the teacher model more stable, thus reducing the influence of inappropriate training epochs, as shown in Fig.~\ref{fig:kl_teacher_experiments_2}.
    
The exponential moving average allows the teacher model to lag behind the student model, and due to the memorization ability of the neural network, the F1-score of the lagged neural network is better than that of the student network at the later stage of training.
The use of consistency loss can promote the output of the student model to approximate the teacher model, so as to alleviate the overfitting problem.
If $\mathcal{L}_2$ is regarded as the consistency loss, it is equivalent to Mean-Teacher~\cite{mean_teacher} being used.
However, according to the results in Table~\ref{tab:KL-Teacher_f1}, $\mathcal{L}_2$ cannot effectively alleviate the overfitting of the student model.
From Table~\ref{tab:KL-Teacher_f1}, better F1-score can be obtained by using $\mathcal{L}_{kl}$ as the consistency loss.
It can be seen from Fig.~\ref{fig:kl_teacher_experiments_2} that the curve of the F1-score using $\mathcal{L}_{kl}$ is at the top, which indicates that \emph{KL-Teacher} alleviates the overfitting phenomenon to some extent through the memorization ability of the neural network.
The analysis of the $\beta$ in KL-Teacher is presented in the Table~\ref{tab:beta}, and the proposed method is robust to $\beta$.
\begin{table}[!h]
\centering
\resizebox{.98\columnwidth}{!}{
\begin{tabular}{cccccccc}
\toprule
$\beta$ & 0           & 0.2         & 0.4         & 0.5         & 0.6         & 0.8         & 1.0         \\ \midrule
F1      & 97.51(0.68) & 97.95(0.38) & 98.20(0.31) & 98.15(0.35) & 98.40(0.31) & 98.55(0.26) & 98.65(0.23) \\ \bottomrule
\end{tabular}}
\caption{Analysis of the $\beta$ in the cotton of HongHu dataset.}
\label{tab:beta}
\end{table}

\subsection*{10. More Experimental Results about Class Prior-based Method with Oracle Class Prior}
The class prior-based method is evaluated with estimated class prior ($\hat{\pi}_p$) and oracle class prior ($\pi_p$).
Due to the severe inter-class similarity and intra-class variation, the $\pi_p$ is hard to be estimated accurately in HSI.
The $\pi_p$ and $\hat{\pi}_p$ are shown in the Table~\ref{tab:class_prior}.
The results of class prior-based method are very poor without accurate $\pi_p$, the proposed method achieves competitive results compared to the class prior-based method with an oracle $\pi_p$ (Table~\ref{tab:class_prior}).
\begin{table}[!h]
  \centering
  \resizebox{.98\columnwidth}{!}{  
  \begin{tabular}{ccccccc}
  \toprule
                               & Class        & Rape                 & Tube mustard         & Cowpea               & Soybean              & Watermelon           \\ \midrule
  \multirow{2}{*}{Class prior} & $\pi_p$          & 0.1317               & 0.0367               & 0.0617               & 0.0279               & 0.0123               \\
                               & $\hat{\pi}_p$          & 0.2231               & 0.3109               & 0.2547               & 0.1509               & 0.3109               \\ \midrule
  \multirow{3}{*}{F1-scores}          & OC Loss($\pi_p$) & \textbf{98.73(0.05)} & 95.97(0.93)          & \textbf{90.43(0.48)} & 98.04(0.97)          & 91.21(1.92)          \\
                               & OC Loss($\hat{\pi}_p$) & 81.81(1.23)          & 23.57(0.22)          & 58.97(3.56)          & 42.34(1.06)          & 12.23(0.46)          \\ \cmidrule{2-7} 
                               & T-HOneCls      & 97.81(0.16)          & \textbf{97.38(0.35)} & 90.31(1.13)          & \textbf{99.13(0.28)} & \textbf{92.99(0.90)} \\ \bottomrule
  \end{tabular}}
  \caption{Comparison of the $\pi_p$ and the $\hat{\pi}_p$.
  Comparison of the F1-scores of class prior-based method with $\pi_p$ and $\hat{\pi}_p$.}
  \label{tab:class_prior}
  \end{table}

\end{document}